\pdfoutput=1
\documentclass{article}
\usepackage{arxiv}
\usepackage[utf8]{inputenc}
\usepackage{subcaption}
\usepackage[T1]{fontenc}
\usepackage{hyperref}
\usepackage{url}
\usepackage{amsmath,amssymb,amsfonts}
\usepackage{algorithmic}
\usepackage{textcomp}
\usepackage{lscape}
\usepackage{gensymb}
\usepackage{longtable}
\usepackage{tabularx}
\usepackage{multicol}
\usepackage{footnote}
\usepackage{wrapfig}
\usepackage{float}
\usepackage{etoolbox}
\robustify\cite
\usepackage{amsmath}
\usepackage{graphicx}
\title{Review of methods for automatic cerebral microbleeds detection}
\author{
    Maria Ferlin \\
  Gdańsk University of Technology \\
  Gdańsk, Poland\\
  \texttt{maria.ferlin@pg.edu.pl} \\
   \And
  Zuzanna Klawikowska \\
  Gdańsk University of Technology \\
  Gdańsk, Poland\\
  \texttt{zuzanna.klawikowska@pg.edu.pl} \\
   \And
    Michał Grochowski \\
  Gdańsk University of Technology \\
  Gdańsk, Poland\\
  \texttt{michal.grochowski@pg.edu.pl} \\
   \And
   Małgorzata Grzywińska \\
  Medical University of Gdańsk \\
  Gdańsk, Poland\\
  \texttt{malgorzata.grzywinska@gumed.edu.pl} \\
   \And
   Edyta Szurowska \\
  Medical University of Gdańsk \\
  Gdańsk, Poland\\
  \texttt{eszurowska@gumed.edu.pl} \\
}
  
\begin{document}
\maketitle

\begin{abstract}
Cerebral microbleeds detection is an important and challenging task. With the gaining popularity of the MRI, the ability to detect cerebral microbleeds also raises. Unfortunately, for radiologists, it is a time-consuming and laborious procedure. For this reason, various solutions to automate this process have been proposed for several years, but none of them is currently used in medical practice. In this context, the need to systematize the existing knowledge and best practices has been recognized as a factor facilitating the imminent synthesis of a real CMBs detection system practically applicable in medicine. To the best of our knowledge, all available publications regarding automatic cerebral microbleeds detection have been gathered, described, and assessed in this paper in order to distinguish the current research state and provide a starting point for future studies.
\end{abstract}

\section{Introduction}
\label{sec:introduction}
\begin{figure}[h!]
    \begin{center}
        \includegraphics[width=0.8\textwidth]{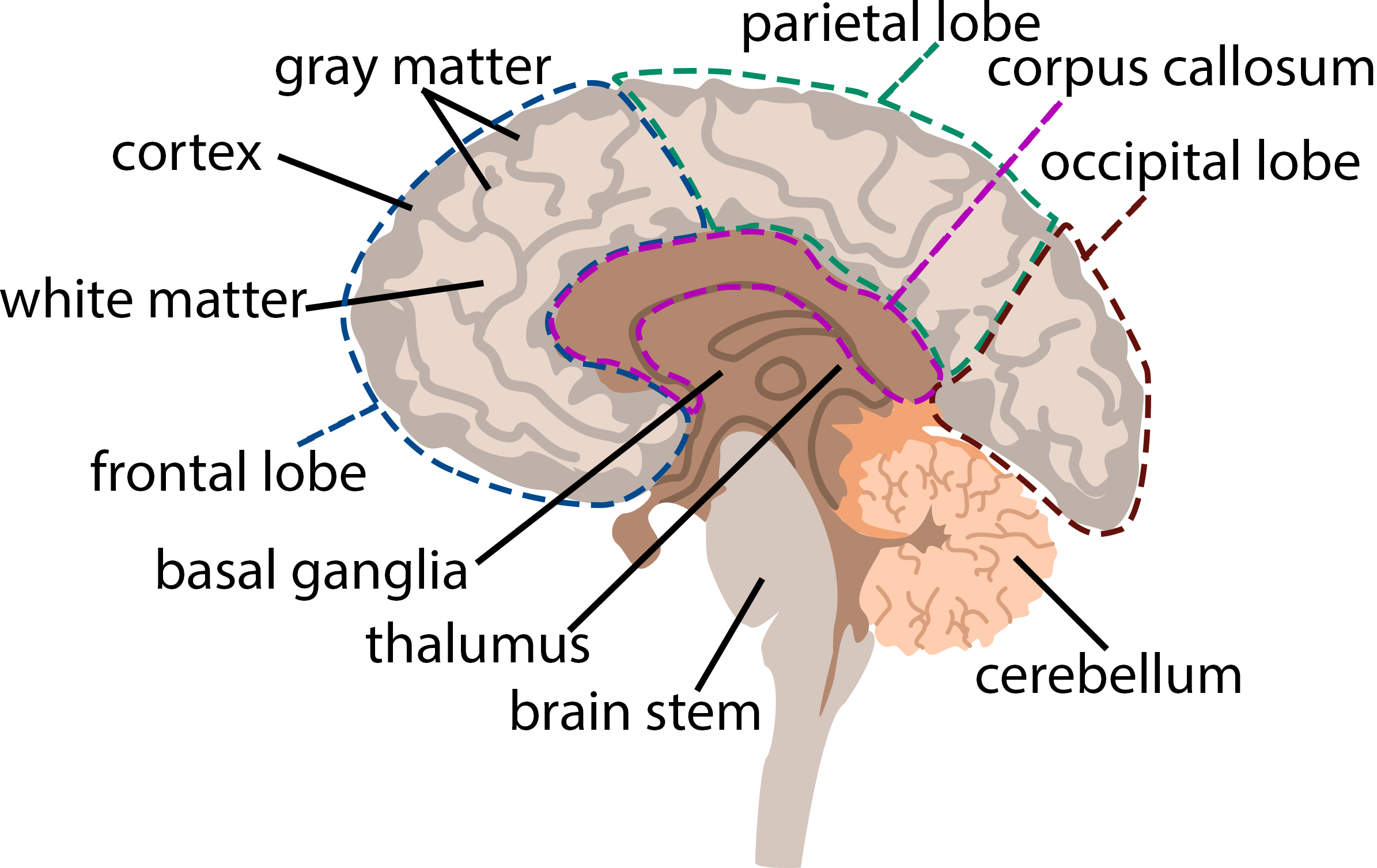}
        \caption{Brain anatomy in the sagittal plane. In addition to the presented structures, the temporal lobe, the insula, and the external and internal capsules, which are not visible in this plane, are also important in the context of scales used to rate CMB. CMBs can be found in all structures indicated in the figure as well as in those mentioned above.
        \label{fig:brain_anatomy}}
    \end{center}
\end{figure}

Cerebral microbleeds (CMBs) are defined as small, homogeneous, hypointense foci well seen on T2*-weighted MRI sequences with the associated so-called ‘blooming effect’.
They are collections of blood degradation products (mainly hemosiderin) that can remain in macrophages for years, following a microhemorrhage\cite{werring_cerebral_2007, shoamanesh_cerebral_2011, martinez-ramirez_cerebral_2014, cordonnier_spontaneous_2007}.
The ‘blooming effect’ takes place when the MRI overestimates the diameter of the microbleed \cite{Greenberg2009CerebralInterpretation}.
CMBs may occur in every region of the brain and can be categorized relative to that area \cite{sveinbjornsdottir_cerebral_2008, Cordonnier2009, gregoire_microbleed_2009}, (Fig. \ref{fig:brain_anatomy}).
They may appear due to a range of pathological processes in the cerebral vessels \cite{martinez-ramirez_cerebral_2014, Shams2016CerebrospinalImpairment., Mazurek}.
Around 5\% of population have microbleeds and they are completely healthy \cite{Akoudad2015CerebralStudy., cordonnier_spontaneous_2007}.
However, the increased number of CMBs in the patient's brain may indicate the existence of some medical condition \cite{Cordonnier2009}.
Additionally, they are sometimes accidentally found in association with other pathologies \cite{bian_susceptibility-weighted_2014}.
Undeniably, however, high prevalence of cerebral microbleeds is closely associated with cognitive disfunction \cite{Yakushiji2008BrainDisorder}.

\begin{figure}
    \begin{center}
        \includegraphics[width=0.8\textwidth]{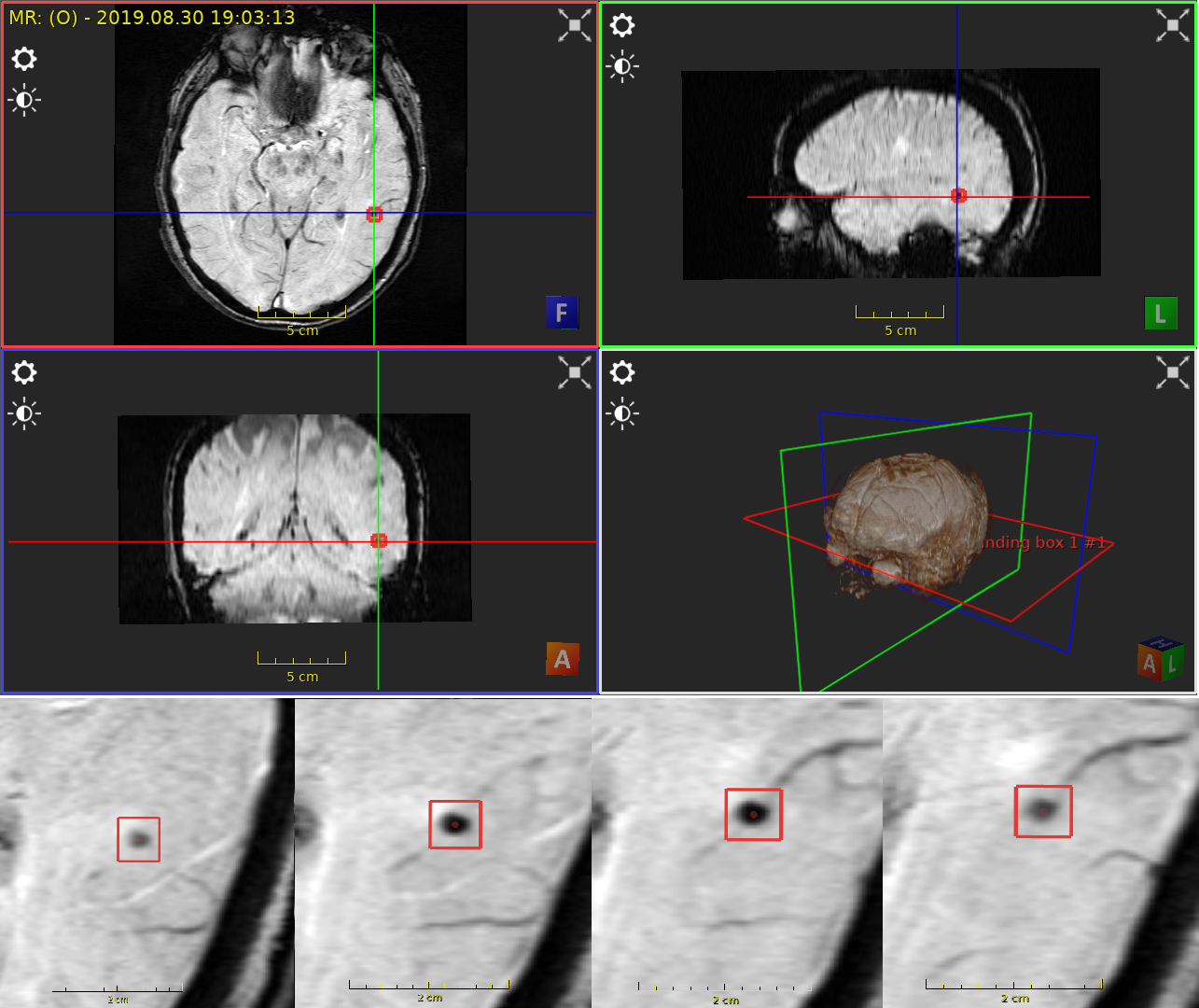}
        \caption{Example of CMB. Upper images present the same microbleed in three planes, while bottom ones present sequence of adjacent slices fragments, in which the microbleed is visible (marked by red frame). Images acquired using ImFusion software.\label{fig:cmb_sample}}
    \end{center}
\end{figure}

CMBs detection is a challenging task due to small size of the lesion compared to the whole image (Fig. \ref{fig:cmb_sample}).
Moreover, there are many lesions that mimic the CMBs.
The main CMB mimics include calcifications, flow voids in pial blood vessels, iron deposits, and deoxyhemoglobin \cite{Greenberg2009CerebralInterpretation}.
Both calcium and iron deposits may appear as small foci of low signal intensity on a T2*-weighted MRI.
Flow voids caught in the cross-sections of cortical sulci can be distinguished from CMBs by their sulcal location, equal visibility on T2-weighted SE and GRE sequences, and linear structure when examined over contiguous slices, particularly evident at smaller slice thickness.
The presence of paramagnetic deoxyhemoglobin in cerebral venules produces its own blooming effect, which requires the rater to rely on their tubular structure for differentiating them from CMBs.
Metastatic melanoma in the brain can appear hypointense on T2*-weighted MRI and may mimic CMB.
Other mimics, such as mineralization of the basal ganglia or diffuse axonal injury, for instance, can be excluded based on the appearance or clinical history.  



CMBs detection is very important considering proper diagnosis and treatment, as they may indicate some major and more complicated issues.
From the medical perspective, the crucial information is the number of detected cerebral microbleeds \cite{haller_2018, shams_swi_2015,Greenberg2009CerebralInterpretation, poels_incidence_2011}. 
Another useful information is their location \cite{werring_cognitive_2004, poels_cerebral_2012, Cordonnier2009, gregoire_microbleed_2009, cordonnier_brain_2011}. Therefore, there is no need to perform  segmentation which is a complex computational algorithm, it is enough to detect them.

With gaining popularity of the MRI as a good imaging tool, the ability to detect cerebral microbleeds also raises.
Unfortunately, for radiologists, it is a time-consuming and laborious process.
Technology and automatic image processing can come to the rescue in this case.


Different solutions have been proposed for last years.
However, the problem is complex and there is no unification and consistency between the researches.
According to our best knowledge, the results achieved so far are still not used in medical practice.
In this context, the authors recognized the need to systematize the existing knowledge and best practices as a factor which will facilitate imminent synthesis of a real CMBs detection system, which would be practically applicable in medicine.

The existing research results are in fact difficult to compare due to various, unavailable publicly datasets and the lack of system evaluation metrics.
The guidelines included in this paper are expected to present new research in a more beneficial way.
Probably, the prevalence of a few publicly available datasets will result in evaluation of new approaches on these datasets.

\subsection{Review criteria}
The aim of this research was to gather all previous works and achievements in the field of cerebral microbleeds detection.
Regarding the lack of order in existing research and comparison ability we decided to collate different approaches and methods, in order to distinguish the current research state and provide a starting point for future studies.
It is noteworthy that the key word in this matter is \textit{automatic} as a guide for a radiologist to detect microbleeds on the MRI existed well beyond \cite{charidimou_cerebral_2011, charidimou_cerebral_2012, charidimou_cerebral_2013, cordonnier_spontaneous_2007, kaaouana_improved_2017, tsushima_mr_2002}.

Firstly, a comprehensive literature review regarding automatic cerebral microbleeds detection have been done.
In order to do that, careful search was performed for all papers connected with this topic in Google Scholar, IEEE Xplore, and Elsevier platforms, using key phrases: \textit{automatic cerebral microbleeds detection}, \textit{automatic CMB detection}, \textit{cerebral microbleeds detection}.

The next step was the search for related papers in the references of all gathered works.
The literature review dates back to year 2011, in which, to the best of our knowledge, first papers about automatic CMB detection were published.

The main information gathered from each paper referred to: database, pre-processing, methods used, proposed approach with the best or the most significant results, conclusions, and challenges.

For the majority of modern methods, the key issue is the availability of datasets, therefore we decided to collate the information about all datasets used in this type of research in Section \ref{sec:data}, which also introduces the issues of MRI and CMB characteristics and CMB rating.

To maintain clarity of the paper, descriptions of particular algorithms are given in Section \ref{sec:methodology}, while the exact approach leveraging from those algorithms is presented in Table \ref{approaches_comparison_table}.
The algorithms described in Section \ref{sec:methodology} are divided into two main groups referring to detection and verification of CMB candidates. Subsection \ref{sec:pre-processing} also presents different pre-processing algorithms that were used to prepare a dataset for training and testing.
Eventually, all methods and algorithms that were used to solve this task are presented.
It turned out during the reported research that the evaluation of results is a challenging problem due to the lack of a standard for the metrics used.
It is not only problematic for existing approaches comparison, but also makes it impossible to assess a specific method itself. To address this, a range of metrics is presented in Subsection \ref{sec:evaluation}, along with their features and dependencies.

Section \ref{sec:discussion} provides a comprehensive assessment of all the presented research, followed by conclusions and challenges, both gathered during literature review and emerging from this analysis.

\section{Data sources}
In order to understand the task of cerebral microbleeds detection it is essential to understand the magnetic resonance imaging, acquisition process and rating procedure.
Therefore, we decided to introduce the process of MR images formation. Further, the relevant sequences and rating scales are described. Finally, we present datasets used for cerebral microbleeds detection.

\label{sec:data}
\subsection{Magnetic Resonance Imaging  Sequences}
Among the types of brain imaging they are CT (computed tomography) and MRI (magnetic resonance imaging). This paper focuses on MRI because it is the most commonly used technique to study CMB. The main reason is the fact that the CT density of the hemorrhage in CMBs rapidly decreases over days as CMBs become indistinguishable with brain tissue after around 7–10 days~\cite{haller_2018}. Consequently, the sensitivity of CT in imaging CMBs is the highest within the first few days of their appearance. On MR images, CMBs remain visible much longer than on CT.

MRI is the imaging technique in which each sequence is a combination of radiofrequency pulses and gradients. There are over a hundred different sequence types, the acronyms of which depend on the manufacturer. Regardless of the type of sequence, the goal is to obtain the signal of a particular tissue - contrast, as quickly as possible - speed, while limiting the artifacts and without altering the signal to noise ratio \cite{noauthor_introduction_nodate}.

\begin{figure}[!ht]
\begin{center}
\includegraphics[width=0.45\textwidth]{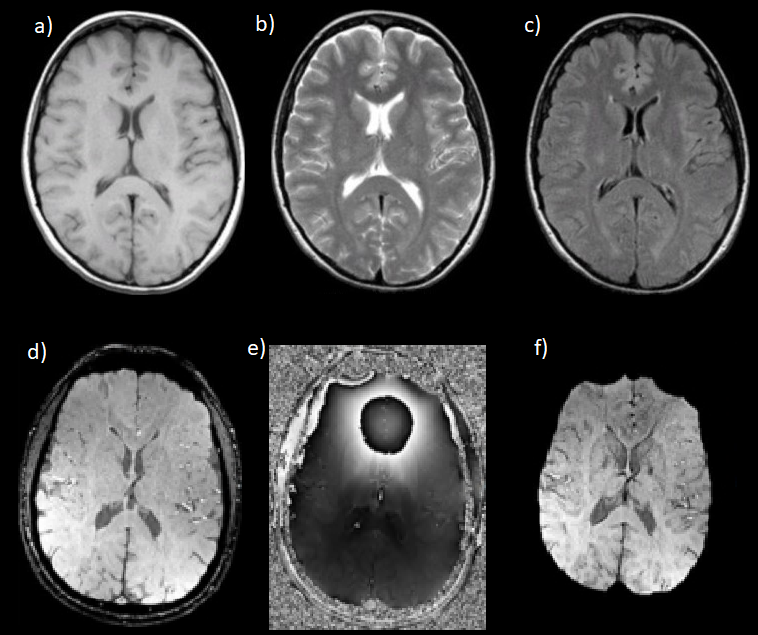}
\caption{Transverse brain plane. Sequences in first row \cite{noauthor_mriBasics_nodate}: T1W (a), T2W (b), FLAIR (c); in second row \cite{liu_susceptibility_2017}: Magnitude (d), Phase (e), SWI (f).\label{fig:sequences}}
\end{center}
\end{figure}

There are three essential components for any imaging sequence.
The first is the radio frequency (RF) excitation pulse which is required for the phenomenon of magnetic resonance.
The second are the gradients for spatial encoding whose arrangement will determine how the k-space is filled.
The third component is signal reading, which combines echo types determining the type of contrast - varying influence of relaxation times: T1, T2 and T2*.
Additionally, more sequence parameters, such as repetition time or flip angle, must be chosen to find a balance between contrast, resolution, and speed \cite{currie_understanding_2013}.

There are three types of relaxation times: T1, T2, and T2* \cite{lipton_image_2008}. The term relaxation means that, once the RF pulse is turned off, the spins are relaxing back into their lowest energy state or to the equilibrium state, realigning with the axis of the magnetic field. T1 is called the longitudinal relaxation time, as it refers to the time needed for the spins to realign along the longitudinal (z)-axis. T2 is defined as the predicted time constant for the decay of transverse magnetization arising from natural interactions at the atomic or molecular level. However, in a real MR experiment, the transverse magnetization decays much faster than would be predicted by natural atomic and molecular mechanisms. This accelerated decay rate is denoted as T2*.



There are two main sequence families, depending on the type of echo recorded. The first family comprises Spin Echo (SE) sequences, which have two essential parameters: TR and TE. They consist of a series of events: 90\degree pulse; 180\degree rephasing pulse at half of echo time (TE) and signal reading at TE, repeated at each time interval TR (Repetition Time). During each repetition, the line of k-space is filled due to different phase encoding.
The example of such sequence is FLuid Attenuation Inversion Recovery (FLAIR). The second family includes Gradient Echo (GE) sequences, during which the flip angle (FA) is usually below 90\degree, which decreases the amount of magnetization tipped into the transverse plane. In this case, there is no 180\degree RF rephasing pulse. The example of this sequence is Susceptibility Weighted Imaging (SWI).
Numerous variations have been developed within each of these families, mainly to increase the acquisition speed.

\begin{figure}
\begin{center}
\includegraphics[width=\columnwidth]{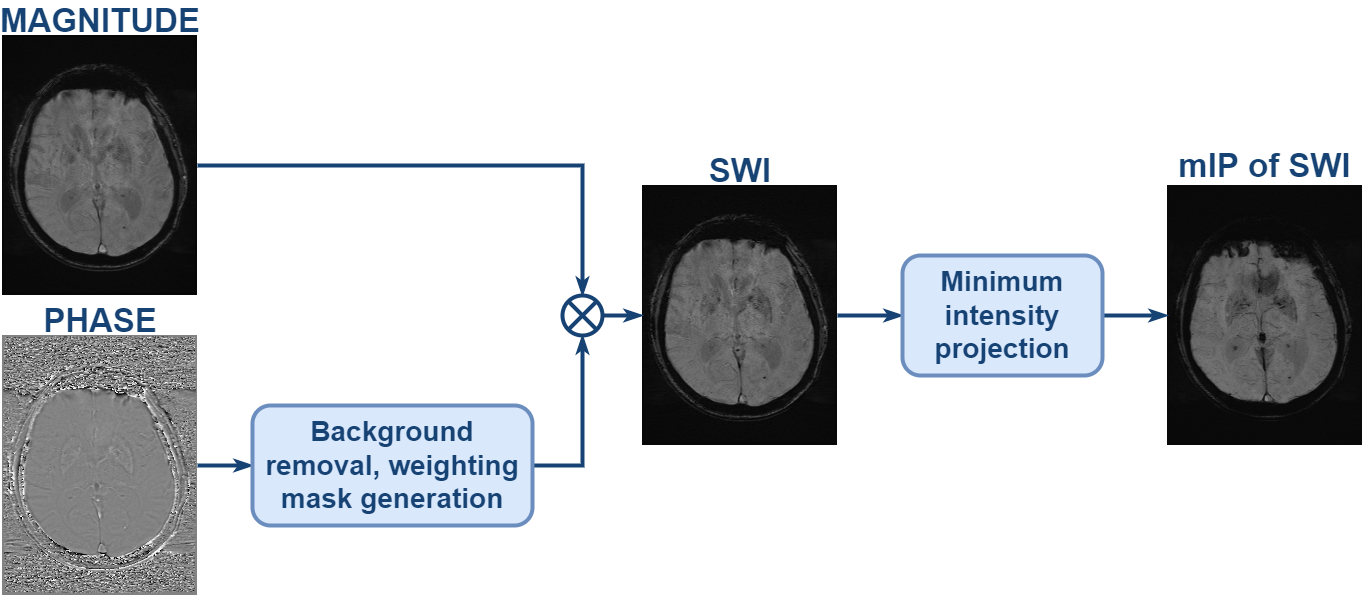}
\caption{Overview of data processing steps in SWI. \label{fig:SWI-steps}}
\end{center}
\end{figure}

A \textbf{T1-weighted} (T1W) sequence demonstrates differences in the T1 relaxation times of tissues. The T1-weighted image is consistent with the anatomy: gray matter is dark and white matter bright. Anatomical gray-white inversion is observed in \textbf{T2-weighted} (T2W) images, in which gray matter is bright and white matter dark. It highlights differences in the T2 relaxation time of tissues. Another sequence is \textbf{FLAIR}, which removes signal from the cerebrospinal fluid (CSF) in the resulting image. Brain tissue in the FLAIR image appears similar to that in the T2W image with gray matter brighter than white matter, but in this case, CSF is dark instead of bright. \textbf{SWI} is a 3D high-spatial-resolution fully velocity corrected gradient-echo MRI sequence which takes advantage of the effect of both phase and magnitude. Fig. \ref{fig:sequences} shows the described sequences and the data processing steps in SWI are shown in Fig. \ref{fig:SWI-steps}.

Susceptibility weighted sequences are named differently depending on the MRI vendor \cite{haller_susceptibility-weighted_2021}. For example, the term SWI is owned by Siemens, GE Healthcare offers a sequence called SWAN , and Philips Healthcare has proposed the name SWIp. Obtaining these sequences differs, due to licensing and patent issues \cite{nandigam_swan_2013}. The differences lie in the use of different ways of combining the sequences, e.g. SWI uses phase and magnitude, while SWAN uses a weighted sum of longer TEs, which preserves T2* dephasing effects, but also increases the signal-to-noise ratio \cite{haller_susceptibility-weighted_2021, hodel_sequences_2012}. However, regardless of the vendor SWI-like sequences are most commonly used in CMB detection, as they have greater sensitivity to this lesion than other sequences \cite{cheng_susceptibility-weighted_2013, akiyama_susceptibility-weighted_2009, nandigam_mr_2009, park_detection_2009, vernooij_cerebral_2008, shams_swi_2015}. It is not only used in terms of automatic detection but also in everyday clinical practice. Another factor that improves the detectability of microbleeds is the strength of the magnetic field \cite{scheid_comparative_2007, conijn_cerebral_2011, de_bresser_visual_2013, nandigam_mr_2009}.

Clinical image data is typically stored in the DICOM format. For scientific analysis, the alternative format is NIFTY.

\subsection{CMB rating}
\label{sec:scales}
Technology that automates clinicians' work should be developed in accordance with clinical practice. It is important to know the ways of assessing a disease, so that the results provided by the proposed tools fit into these guidelines. Two ways used by clinicians to assess CMB are \textbf{Brain Observer Microbleed Scale} (BOMBS) \cite{Cordonnier2009} and \textbf{Microbleed Anatomical Rating Scale} (MARS) \cite{gregoire_microbleed_2009}, proposed in 2009. The evaluation categories are presented in Table \ref{scales_comparison_table}.
Standardized CMB rating scales provide a uniform assessment methodology and enable easy and reliable quantification and categorization of CMBs even when the scales are used by observers with different backgrounds or experience, and thus increase the reliability of the measurement.

Measurement reliability refers to the consistency or repeatability of the measurement. Low reliability indicates large differences in measurement while retesting.
It precludes reproduction or interpretation of the results, and finally makes distinguishing between participants with and without specific medical conditions impossible due to significant measurement error. In clinical evaluation, a measurement error can be introduced by the observer. Therefore, determining observer (clinician) reliability is important for making full comparison of measurement reliability between studies. There are two ways of doing it -- inter- and intra-observer agreement.

\textbf{Intra-observer} agreement determines the degree of agreement between the two studies that use the same technique, in the same patient, obtained by the one observer \cite{filippi_intra-_1995}. \textbf{Inter-observer} agreement determines the degree of agreement between the two studies that use the same technique, in the same patient, obtained by the two observers \cite{filippi_intra-_1995}.


The reliable rating of CMBs presence, number, and location is the important factor for further diagnosis of various diseases. However, many research institutions use their own methods to rate CMBs. Although their reliability based on intra- and inter-observer agreement is reported, details of the methods used are usually not described \cite{charidimou_cerebral_2012}. 


\begin{table}
\caption{CMBs evaluation categories according to rating scales}
\label{scales_comparison_table}
    \begin{tabular}{m{0.45\columnwidth}|m{0.45\columnwidth}}
        \hfil \textbf{BOMBS} &  \hfil \textbf{MARS}\\
        \hline
        \begin{itemize}
            \item certainty:
                \begin{enumerate}
                    \item certain,
                    \item uncertain,
                \end{enumerate}
            \item size:
                \begin{enumerate}
                    \item $<$5 mm,
                    \item 5-10 mm,
                \end{enumerate}
            \item side of brain:
                \begin{enumerate}
                    \item left,
                    \item right,
                \end{enumerate}
            \item location (Fig. \ref{fig:brain_anatomy}):
                \begin{enumerate}
                    \item lobar:
                    \begin{enumerate}
                        \item cortex/gray–white junction,
                        \item \mbox{subcortical white} matter,
                    \end{enumerate}
                    \item deep:
                    \begin{enumerate}
                        \item basal ganglia,
                        \item \mbox{internal and} \mbox{external} \mbox{capsules},
                        \item thalamus,
                    \end{enumerate}
                    \item posterior fossa:
                    \begin{enumerate}
                        \item brain stem,
                        \item cerebellum.
                        \item[]
                    \end{enumerate}
                \end{enumerate}
        \end{itemize} 
        &
        \begin{itemize}
            \item appearance of the lesion:
            \begin{enumerate}
                \item definite,
                \item possible,
            \end{enumerate}
            \item side of brain:
            \begin{enumerate}
                \item left,
                \item right,
            \end{enumerate}
            \item location (Fig. \ref{fig:brain_anatomy}):
            \begin{enumerate}
                \item lobar:
                \begin{enumerate}
                    \item frontal,
                    \item parietal,
                    \item temporal,
                    \item occipital,
                    \item insula,
                \end{enumerate}
                \item deep:
                \begin{enumerate}
                    \item basal ganglia,
                    \item internal capsule,
                    \item external capsule,
                    \item thalumus,
                    \item corpus callosum,
                    \item \mbox{deep and} \mbox{periventricular} white matter,
                \end{enumerate}
                \item infratentorial:
                \begin{enumerate}
                    \item brain stem,
                    \item cerebellum.
                \end{enumerate}
            \end{enumerate}
        \end{itemize}
    \end{tabular}
\end{table}

\begin{landscape}
\scriptsize
\begin{longtable}{p{0.7cm} p{0.7cm} p{1.8cm} p{1.5cm} c p{1cm} p{1cm} p{1.5cm} p{1.2cm} p{1.7cm} p{1.3cm} p{0.9cm} p{0.8cm} p{0.8cm}}

\caption{Comparison of dataset acquisition parameters used in the reviewed approaches.}
\label{datasets_comparison_table}
\endfirsthead
\caption{\textbf{Continued:} Datasets acquisition parameters comparison}
\endhead

    \hline
    \hfil ref. &
    \begin{tabular}[c]{@{}c@{}}\# of\\subject\\/\# of CMB\end{tabular} &
    \hfil RES [mm$^2$] &
    \hfil ST [mm] &
    \hfil TR [ms] &
    \hfil TE [ms] &
    \hfil FA [\degree] &
    \hfil \begin{tabular}[c]{@{}c@{}}BW\\{[}Hz/px]\end{tabular} &
    IMS [voxels] &
    \begin{tabular}[c]{@{}c@{}} FOV\\  $[$ mm$^3$\textbackslash mm$^2$\textbackslash mm $]$ \end{tabular} &
    \hfil Sequences &
    $\beta$~[T] &
    \hfil Rating &
    \hfil Avail.\\ 
    
    \hline
    
    \hfil \cite{kuijf_detecting_2011}  &
    \hfil 2 / 4&
    \hfil 0.35x0.35 &
    \hfil 0.3&
    \hfil 20 &
    \hfil2.5/15&
    \hfil - &
    \hfil - &
    \hfil - &
    \hfil - &
    \hfil T2*W &
    \hfil 7 &
    \hfil MARS &
    \hfil \begin{tabular}[c]{@{}c@{}}on\\request\end{tabular}\\

    \hline
    
    \hfil \cite{barnes_semiautomated_2011} &
    \hfil 6 / 26 &
    \hfil 0.5x1 &
    \hfil 2 &
    \hfil 57 &
    \hfil 40 &
    \hfil 20 &
    \hfil - &
    \hfil 512x320x48 &
    \hfil -&
    \hfil \begin{tabular}[c]{@{}c@{}}Fully Flow-\\Compensated\\3 DGRE\end{tabular} &
    \hfil 1.5 &
    \hfil \cite{ayaz_imaging_2010} &
    \hfil -\\
    
    \hline
    \cite{hong_detecting_2019, hong_improvement_2019, hong_classification_2020} &
    \hfil 10 / - &
    \hfil 0.5x0.5 &
    \hfil 2 &
    \hfil - &
    \hfil 20 &
    \hfil 15 &
    \hfil 120 &
    \hfil 364x448x48 &
    \hfil -&
    \hfil SWI &
    \hfil 3 &
    \hfil MARS &
    \hfil -\\
    

    
    \hline
    
    
    
    \cite{bian_computer-aided_2013,morrison_user-guided_2018} &
    \hfil 15~/~420 &
    \hfil 0.5x0.5 &
    \hfil 2 &
    \hfil 56 &
    \hfil 28 &
    \hfil 20 &
    \hfil - &
    \hfil u x u 40 &
    \hfil 240&
    \hfil T2*W &
    \hfil 3 &
    \hfil \begin{tabular}[c]{@{}c@{}}similar to\\ BOMBS/\\MARS\end{tabular} &
    \begin{tabular}[c]{@{}c@{}}10\\subjects\end{tabular}\\
    
    \hline
    
    \hfil \cite{kuijf_efficient_2012} &
    \hfil 18 / 54&
    \begin{tabular}[c]{@{}c@{}}0.35x0.35 T2*W,\\0.66x0.66 T1W\end{tabular} &
    \hfil \begin{tabular}[c]{@{}c@{}}0.3 T2*W,\\0.7 T1W\end{tabular} &
    \begin{tabular}[c]{@{}c@{}}20 T2*W,\\7 T1W\end{tabular} &
    \hfil \begin{tabular}[c]{@{}c@{}}2.5/15 T2*W,\\3 T1W\end{tabular} &
    \hfil - &
    \hfil - &
    \hfil - &
    \hfil - &
    \hfil \begin{tabular}[c]{@{}c@{}}T2*W,\\ T1W turbo\\field echo\end{tabular} &
    \hfil 7 &
    \hfil MARS &
    \begin{tabular}[c]{@{}c@{}}on\\request\end{tabular}\\
    
    \hline
    
    \begin{tabular}[c]{@{}c@{}}\cite{lu_acerebral_2021,lu_cerebral_2021},\\\cite{zhang_sparse_2016, zhang_voxelwise_2018, zhang_seven-layer_2018},\\ \cite{wang_cerebral_2019, wang_cerebral_2017}\end{tabular} &
    \hfil 20 / - &
    \hfil 0.5x0.5 &
    \hfil 2 &
    \hfil 28 &
    \hfil 20 &
    \hfil 15 &
    \hfil 120 &
    \hfil 364x448×48 &
    \hfil -&
    \hfil SWI &
    \hfil 3 &
    \hfil MARS &
    \hfil -\\
    
    \hline

    \hfil \cite{afzal_transfer_2022} &
    \hfil 20 / -&
    \hfil 0.45x0.45 &
    \hfil 2 &
    \hfil 17 &
    \hfil 24 &
    \hfil - &
    \hfil - &
    \hfil - &
    \hfil -&
    \hfil SWI &
    \hfil 3 &
    \hfil - &
    \hfil -\\
    
    \hline

    \begin{tabular}[c]{@{}c@{}}\cite{chen_automatic_2015, ateeq_ensemble-classifiers-assisted_2018},\\\cite{electronics10182208, liu_cerebral_2020} \\ \cite{qi_dou_automatic_2015} \\ \cite{dou_automatic_2016}\end{tabular} &
    \begin{tabular}[c]{@{}c@{}} 20 / 117,\\44 / 615,\\320 / 1149 \end{tabular} &
    \hfil 0.45x0.45 &
    \hfil 2 &
    \hfil 17 &
    \hfil 24 &
    \hfil - &
    \hfil - &
    \hfil 512x512x150 &
    \hfil 230x230&
    \hfil SWI &
    \hfil 3 &
    \begin{tabular}[c]{@{}c@{}}-,\\MARS,\\MARS \end{tabular}&
    \begin{tabular}[c]{@{}c@{}}20\\subjects\end{tabular}\\
    
    \hline
    
    \hfil \cite{rashid_deepmir_2021} &
    24~/~$>$157 &
    \hfil1x1 &
    \begin{tabular}[c]{@{}c@{}}1\\ T1WMP \& T2W \\ 1.5 SWI\end{tabular} &
    \begin{tabular}[c]{@{}c@{}}1900\\T1W MPRAGE\\(T1WMP),\\3200 T2W,\\35 SWI\end{tabular} &
    \begin{tabular}[c]{@{}c@{}}2.93\\T1WMP,\\408 T2W,\\7.5/\\15/\\22.5/\\30\\SWI\end{tabular} &
    \begin{tabular}[c]{@{}c@{}}9\\T1WMP,\\120 T2W,\\ 15 SWI\end{tabular} &
    \begin{tabular}[c]{@{}c@{}}170\\T1WMP,\\750 T2W,\\200 SWI\end{tabular} &
    \begin{tabular}[c]{@{}c@{}}256x256x176 \\ T1WMP \& T2W,\\ 256x192x96 SWI\end{tabular} &
    \hfil -&
    \hfil \begin{tabular}[c]{@{}c@{}}T1WMP,\\T2W,\\SWI \end{tabular}&
    \hfil 3 &
    Inspired by BOMBS &
    \begin{tabular}[c]{@{}c@{}}on\\request\end{tabular}\\
    
    \hline

    \hfil \cite{sa-ngiem_cerebral_2019} &
    \hfil 26 / - &
    \hfil - &
    \hfil 3 &
    \hfil - &
    \hfil - &
    \hfil - &
    \hfil - &
    u x u x 40-60 &
    \hfil -&
    \hfil SWI &
    \hfil - &
    \hfil - &
    \hfil -\\
    
    \hline
    
    \hfil \cite{ourselin_cerebral_2015} &
    \hfil 26~/~404 &
    \begin{tabular}[c]{@{}c@{}}0.45x0.45 SWI,\\1x1 T1W-MPRAGE\\(magnetization\\prepared\\rapid\\gradient\\ echo)\end{tabular} &
    \begin{tabular}[c]{@{}c@{}}2 SWI,\\1 T1W-MPRAGE\end{tabular} &
    \hfil - &
    \hfil 25 &
    \hfil - &
    \hfil - &
    \hfil - &
    \hfil -&
    \hfil \begin{tabular}[c]{@{}c@{}}SWI,\\T1W-MPRAGE \end{tabular} &
    \hfil 3 &
    \hfil - &
    \hfil -\\
    
    \hline
    
    \hfil \begin{tabular}[c]{@{}c@{}}\cite{fazlollahi_automatic_2013} \\ \cite{fazlollahi_efficient_2014} \\ \cite{fazlollahi_computer-aided_2015}\end{tabular} &
    \hfil \begin{tabular}[c]{@{}c@{}}30 / 64,\\41 / 103,\\66 / 231\end{tabular} &
    \begin{tabular}[c]{@{}c@{}}0.93x0.93 SWI,\\1x1 T1W\end{tabular} &
    \hfil \begin{tabular}[c]{@{}c@{}}1.75 SWI,\\1.2 T1W\end{tabular}&
    \begin{tabular}[c]{@{}c@{}}27 SWI,\\2.3 T1W\end{tabular} &
    \begin{tabular}[c]{@{}c@{}}20 SWI,\\2.98 T1W\end{tabular} &
    \begin{tabular}[c]{@{}c@{}}20 SWI,\\9 T1W\end{tabular} &
    \hfil - &
    \hfil - &
    \hfil \begin{tabular}[c]{@{}c@{}}240x256 T1W,\end{tabular}&
    \hfil SWI, T1W &
    \hfil 3 &
    \hfil \begin{tabular}[c]{@{}c@{}}-,\\MARS,\\MARS\end{tabular} &
    \begin{tabular}[c]{@{}c@{}}on\\request\end{tabular}\\
    
    \hline
    
    \cite{van_den_heuvel_automated_2016, t_l__a_van_den_heuvel_computer_2015}&
    \hfil 51~/~627 &
    \begin{tabular}[c]{@{}c@{}}0.98x0.98 SWI,\\1x1 T1 MP-RAGE\\(T1MPR)\end{tabular} &
    \hfil - &
    \begin{tabular}[c]{@{}c@{}}27 SWI,\\2300 T1MPR\end{tabular} &
    \begin{tabular}[c]{@{}c@{}}20 SWI,\\2.98 T1MPR\end{tabular} &
    \begin{tabular}[c]{@{}c@{}}15 SWI,\\9 T1MPR\end{tabular} &
    \begin{tabular}[c]{@{}c@{}}120 SWI,\\240T1 MPR\end{tabular} &
    \hfil - &
    \hfil -&
    \hfil \begin{tabular}[c]{@{}c@{}} SWI,\\T1-MPRAGE \end{tabular}&
    \hfil 3 &
    \hfil MARS &
    \hfil -\\
    
    \hline

    \hfil \cite{li_detecting_2021} &
    \hfil 58~/~1301&
    \hfil - &
    \begin{tabular}[c]{@{}c@{}}5\\T2F \& T2WF,\\ 2 SWAN-W\end{tabular} &
    \begin{tabular}[c]{@{}c@{}}5727\\T2 FRFSE (T2F),\\ 77.3 SWAN-W,\\ 8400\\T2W FLAIR (T2WF)\end{tabular} &
    \begin{tabular}[c]{@{}c@{}}93 T2F,\\ 45 SWAN-W,\\ 145 T2WF\end{tabular} &
    \begin{tabular}[c]{@{}c@{}}15\\SWAN-W,\\ 145\\T2WF \end{tabular} &
    \begin{tabular}[c]{@{}c@{}}833\\T2F \& T2WF,\\ 625\\SWAN-W \end{tabular} &
    \hfil 512x512 x u &
    \hfil 240&
    \hfil \begin{tabular}[c]{@{}c@{}}T2F,\\SWAN-W,\\ T2WF\end{tabular}&
    \hfil 3 &
    \hfil - &
    \hfil -\\
    
    \hline
    
    \hfil \cite{chesebro_automated_2021} &
    \hfil 72 / 64 &
    \begin{tabular}[c]{@{}c@{}}0.43×0.43\\T2*W SWI,\\0.43×0.43\\ T2*W GRE\end{tabular} &
    \begin{tabular}[c]{@{}c@{}}1 T1W,\\2 T2*W SWI,\\1 T2*W GRE\end{tabular} &
    \begin{tabular}[c]{@{}c@{}}6.6 T1W,\\17 T2*W SWI,\\15 T2*W GRE\end{tabular} &
    \begin{tabular}[c]{@{}c@{}}3 T1W,\\24 T2*W SWI,\\22 T2*W GRE\end{tabular} &
    \hfil - &
    \hfil - &
    \hfil - &
    \begin{tabular}[c]{@{}c@{}}256x200 T1W,\\244x197 T2*W SWI,\\220x181 T2*W GRE\end{tabular}&
    \hfil \begin{tabular}[c]{@{}c@{}}T1W,\\T2*W SWI,\\T2*W\\ GRE\end{tabular} &
    \hfil 3 &
    \hfil \cite{Greenberg2009CerebralInterpretation} &
    \begin{tabular}[c]{@{}c@{}}on\\request\end{tabular}\\
    
    \hline
    
    \hfil \cite{kuijf_semi-automated_2013} &
    \hfil 72~/~148 &
    \begin{tabular}[c]{@{}c@{}}0.96x0.96\\T2*W \& FLAIR,\\1x1 T1W\end{tabular} &
    \begin{tabular}[c]{@{}c@{}}3 T2*W \&\\ FLAIR,\\ 1 T1W\end{tabular} &
    \begin{tabular}[c]{@{}c@{}}1653 T2*W,\\ 11000 FLAIR,\\ 7.9 T1W\end{tabular} &
    \begin{tabular}[c]{@{}c@{}}20 T2*W,\\ 125 FLAIR,\\ 4.5 T1W\end{tabular} &
    \hfil - &
    \hfil - &
    \hfil - &
    \hfil -&
    \hfil \begin{tabular}[c]{@{}c@{}} T2*W,\\FLAIR,\\T1W turbo\\field echo \end{tabular}&
    \hfil 3 &
    \hfil MARS &
    \begin{tabular}[c]{@{}c@{}}on\\request\end{tabular}\\
    
    \hline
    
    \begin{tabular}[c]{@{}c@{}}\cite{al-masni_automated_2020, al-masni_two_2020},\\\cite{electronics10182208}\end{tabular}&
    \begin{tabular}[c]{@{}c@{}}72 / 188\\HR,\\107 / 572\\ LR\end{tabular} &
    \hfil \begin{tabular}[c]{@{}c@{}}0.50×0.50\\HR,\\0.80x0.80\\ LR\end{tabular} &
    \hfil 2  &
    \begin{tabular}[c]{@{}c@{}}27 HR,\\40 LR\end{tabular} &
    \begin{tabular}[c]{@{}c@{}}20 HR,\\13.7 LR\end{tabular} &
    \hfil 15 &
    \hfil 120 &
    \begin{tabular}[c]{@{}c@{}}512x448x72 HR,\\288x252x72 LR\end{tabular} &
    \hfil \begin{tabular}[c]{@{}c@{}}256x224 HR,\\201×229 LR\end{tabular} &
    \hfil \begin{tabular}[c]{@{}c@{}} SWI,\\Phase,\\ Magnitude\end{tabular}&
    \hfil 3 &
    \hfil \cite{Greenberg2009CerebralInterpretation} &
    \hfil no\\
    
    \hline
    
    \hfil \cite{chen_toward_2019} &
    \hfil 73~/~2835 &
    \hfil 0.5x0.5 &
    \hfil \begin{tabular}[c]{@{}c@{}}1 SWI,\\ 2\\3DSPGR\end{tabular} &
    \begin{tabular}[c]{@{}c@{}}40 SWI,\\ 50\\3DSPGR\end{tabular} &
    \begin{tabular}[c]{@{}c@{}}2.4/12/14.3/20.3 \\SWI,\\ 16\\3DSPGR\end{tabular} &
    \hfil 25 &
    \hfil - &
    \hfil 512x512 x u &
    \hfil - &
    \hfil \begin{tabular}[c]{@{}c@{}}4-echo\\3D TOF-SWI,\\ 3DSPGR\\ (3D Spoiled\\Gradient\\Recalled)\end{tabular} &
    \hfil 7 &
    \begin{tabular}[c]{@{}c@{}}computer-\\aided\\detection\\ developed\\by \cite{bian_computer-aided_2013}\\with rater\end{tabular} &
    \hfil -\\
    
    \hline
    
    \hfil \cite{seghier_microbleed_2011} &
    \hfil 74 / - &
    \hfil 0.938x0.938 &
    \hfil 5 &
    \begin{tabular}[c]{@{}c@{}}6000\\T2W Fast Spin Echo\\ (T2WFSE),\\300 T2*GRE\end{tabular} &
    \begin{tabular}[c]{@{}c@{}}105 T2WFSE,\\40 T2*GRE\end{tabular} &
    \begin{tabular}[c]{@{}c@{}}20\\T2*GRE \end{tabular}&
    \hfil - &
    \hfil \begin{tabular}[c]{@{}c@{}} 256x224x u\\T2WFSE\end{tabular} &
    \hfil 240x180 &
    \begin{tabular}[c]{@{}c@{}}T2WFSE,\\T2* GRE\end{tabular} &
    \hfil 1.5 &
    \hfil MARS &
    \begin{tabular}[c]{@{}c@{}}on\\request\end{tabular}\\
    
    \hline
 
    \hfil \cite{Myung2021NovelPerformance} &
    186~/~1716 &
    \hfil 0.63x0.63 &
    \hfil 2 &
    \hfil 1050 &
    \hfil 20 &
    \hfil 21 &
    \hfil - &
    \hfil - &
    \hfil 220x198&
    3D Fast Field-Echo &
    \hfil 3 &
    \hfil - &
    \hfil -\\
    
    \hline

    \hfil \cite{momeni_synthetic_2021} &
    214~/~235 &
    \begin{tabular}[c]{@{}c@{}}0.93×0.93 SWI,\\1x1 T1W \end{tabular}&
    \hfil \begin{tabular}[c]{@{}c@{}}1.75 SWI,\\1.2 T1W\end{tabular}&
    \hfil \begin{tabular}[c]{@{}c@{}} 27 SWI,\\2.3 T1W \end{tabular}&
    \begin{tabular}[c]{@{}c@{}}20 SWI,\\2.98 T1W\end{tabular}&
    \hfil \begin{tabular}[c]{@{}c@{}}20 SWI,\\9 T1W\end{tabular}&
    \hfil - &
    \begin{tabular}[c]{@{}c@{}}u x u x 160\\T1W\end{tabular} &
    \hfil 240x256 T1W&
    \hfil SWI, T1W &
    \hfil 3 &
    \hfil MARS &
    \begin{tabular}[c]{@{}c@{}}on\\request\end{tabular}\\
    
    \hline
   
    \hfil \cite{liu_cerebral_2019} &
    220~/~1011 &
    \begin{tabular}[c]{@{}c@{}}0.45-0.53x\\0.57-1.05 1.5T,\\0.50-0.54x\\0.50-1.07 3T\end{tabular} &
    \begin{tabular}[c]{@{}c@{}}2-2.65 1.5T,\\2/2.3 3T\end{tabular}&
    \begin{tabular}[c]{@{}c@{}}49/50 1.5T,\\27-34 3T\end{tabular} &
    \begin{tabular}[c]{@{}c@{}}40 1.5T,\\17.5-20 3T\end{tabular} &
    \begin{tabular}[c]{@{}c@{}}15 1.5T,\\12/15 3T\end{tabular} &
    \begin{tabular}[c]{@{}c@{}}80 1.5T,\\100-425 3T\end{tabular} &
    \hfil \begin{tabular}[c]{@{}c@{}}512x\\304-448x\\56/60 1.5T,\\448-512x\\322-416x\\56/128 3T\end{tabular} &
    \hfil - &
    \hfil - &
    \hfil 1.5/3 &
    \hfil - &
    \begin{tabular}[c]{@{}c@{}}on\\request\end{tabular}\\
    
    \hline
    
    \hfil \cite{ghafaryasl_computer_2012} &
    237~/~631&
    \hfil 0.5x0.5 &
    \hfil \begin{tabular}[c]{@{}c@{}}1.6 T2*W,\\0.8 GRE PDW\end{tabular} &
    - &
    \hfil - &
    \hfil - &
    \hfil - &
    \hfil - &
    \hfil - &
    \begin{tabular}[c]{@{}c@{}}3D T2*W,\\GRE\\Proton-\\Density\\ Weighted \end{tabular}&
    \hfil 1.5 &
    \hfil - &
    \hfil -\\
    
    \hline
    
    \hfil \cite{sundaresan_automated_2022} &
    \begin{tabular}[c]{@{}c@{}}270 / $>$505\end{tabular} &
    \hfil \begin{tabular}[c]{@{}c@{}}0.9x0x8\\T2*-GRE,\\0.8x0.8\\ SWI\end{tabular} &
    \begin{tabular}[c]{@{}c@{}}5 T2*-GRE,\\3 SWI\end{tabular} &
    \begin{tabular}[c]{@{}c@{}}504 T2*-GRE,\\27 SWI\end{tabular} &
    \begin{tabular}[c]{@{}c@{}}15 T*2-GRE,\\9.4/20 SWI\end{tabular} &
    \hfil - &
    \hfil - &
    \begin{tabular}[c]{@{}c@{}}640x640x28\\T2*-GRE,\\256x288x48 SWI\end{tabular} &
    \hfil -&
    T2*-GRE, SWI &
    \hfil -/3 &
    \hfil MARS &
    \begin{tabular}[c]{@{}c@{}}on\\request\end{tabular}\\
    
    \hline

    \hfil \cite{stanley_automated_2022} &
    \begin{tabular}[c]{@{}c@{}}320 / 114\\ SWI,\\179 / 760\\ SVS\end{tabular} &
    \hfil \begin{tabular}[c]{@{}c@{}}0.45x0.45\\SWI\end{tabular}&
    \hfil 2 &
    \begin{tabular}[c]{@{}c@{}}17 SWI,\\27/40 SVS\end{tabular} &
    \hfil \begin{tabular}[c]{@{}c@{}}24 SWI,\\20/14 SVS\end{tabular} &
    \hfil \begin{tabular}[c]{@{}c@{}}15\\SVS\end{tabular}&
    \hfil 120 SVS &
    \begin{tabular}[c]{@{}c@{}}512x512x150\\SWI\end{tabular} &
    \hfil 230x230 SWI &
    SWI &
    \hfil 3 &
    \hfil \begin{tabular}[c]{@{}c@{}} MARS\\SWI /\\ \cite{Greenberg2009CerebralInterpretation}\\SVS \end{tabular}&
    \hfil -\\
    
    \hline


\end{longtable}
\end{landscape}

\begin{figure*}
\centering
    \includegraphics[width=\textwidth]{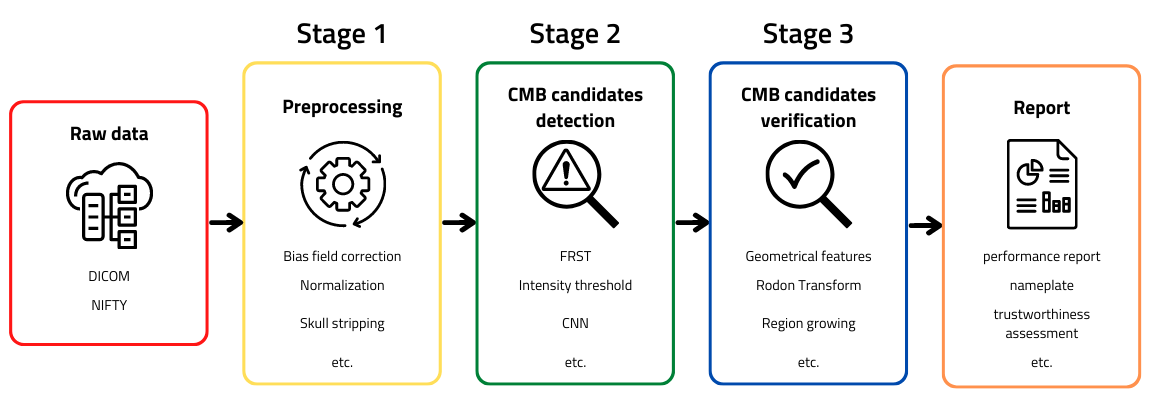}
    \caption{Pipeline of a typical CMB detection approach. \label{fig:pipeline}}
\end{figure*}

\subsection{Datasets}
Studies on the tools that automatically detect microbleeds can be divided into three categories based on the source of the data used. The first category includes researches that use data from specific, existing studies: \cite{barnes_semiautomated_2011, chesebro_automated_2021, fazlollahi_automatic_2013, fazlollahi_efficient_2014, fazlollahi_computer-aided_2015, ghafaryasl_computer_2012, gunter_p4-232_2018, kuijf_detecting_2011, kuijf_efficient_2012, kuijf_semi-automated_2013, liu_cerebral_2019, momeni_synthetic_2021, nandigam_mr_2009, rashid_deepmir_2021, sundaresan_automated_2022}. The second category refers to studies that collected data specifically to develop this tools: \cite{al-masni_automated_2020, al-masni_two_2020, bian_computer-aided_2013, morrison_user-guided_2018, chen_toward_2019, chen_automatic_2015, qi_dou_automatic_2015, dou_automatic_2016, chen_chapter_2017, dou_chapter_2020, ateeq_ensemble-classifiers-assisted_2018, electronics10182208, hong_detecting_2019, hong_improvement_2019, hong_classification_2020, kuijf_semi-automated_2013, li_detecting_2021, liu_cerebral_2012, liu_cerebral_2020, lu_detection_2017, lu_acerebral_2021, lu_cerebral_2021, seghier_microbleed_2011, stanley_automated_2022, zhang_sparse_2016, zhang_voxelwise_2018, zhang_seven-layer_2018}, whereas the third category contains studies that do not specify the sources of data used: \cite{bao_voxelwise_2018, t_l__a_van_den_heuvel_computer_2015, van_den_heuvel_automated_2016, Myung2021NovelPerformance, ourselin_cerebral_2015, wang_cerebral_2017, wang_cerebral_2019, wang_cerebral_2020, sa-ngiem_cerebral_2019, doke_using_2020, lu_diagnosis_2020, afzal_transfer_2022, tajudin_microbleeds_2017, k_standvoss_cerebral_2018, tao_voxelwise_2018}.
The researches focused on diagnosing several medical conditions: Alzheimer's and elderly diseases (AD) \cite{barnes_semiautomated_2011, chesebro_automated_2021, fazlollahi_automatic_2013, fazlollahi_efficient_2014, fazlollahi_computer-aided_2015, ghafaryasl_computer_2012, gunter_p4-232_2018, kuijf_semi-automated_2013, momeni_synthetic_2021},
Cerebral Autosomal Dominant Arteriopathy with Subcortical Infarcts and Leukoencephalopathy (CADASIL) \cite{bao_voxelwise_2018, hong_detecting_2019, hong_improvement_2019, hong_classification_2020, lu_acerebral_2021, lu_cerebral_2021, wang_cerebral_2017, wang_cerebral_2019, wang_cerebral_2020, zhang_sparse_2016, zhang_voxelwise_2018, zhang_seven-layer_2018, doke_using_2020, lu_diagnosis_2020, tao_voxelwise_2018}, 
Second Manifestation of Arterial Disease (SMART) \cite{kuijf_detecting_2011, kuijf_efficient_2012},
Traumatic Brain Injury (TBI) \cite{t_l__a_van_den_heuvel_computer_2015, van_den_heuvel_automated_2016, liu_cerebral_2019, ourselin_cerebral_2015, k_standvoss_cerebral_2018}, 
stroke \cite{liu_cerebral_2012, liu_cerebral_2019, seghier_microbleed_2011, sundaresan_automated_2022}, 
Intracerebral Haemorrhages (ICH) \cite{nandigam_mr_2009, sundaresan_automated_2022}, 
gliomas \cite{bian_computer-aided_2013, morrison_user-guided_2018, chen_toward_2019}, 
hemodialysis cases \cite{liu_cerebral_2019}, 
Cerebral Amyloid Angiopathy (CAA) \cite{nandigam_mr_2009}, 
atherosclerosis \cite{rashid_deepmir_2021}, 
or did not distinguish any particular disease besides the appearance of CMBs \cite{al-masni_automated_2020, al-masni_two_2020, chen_automatic_2015, qi_dou_automatic_2015, dou_automatic_2016, chen_chapter_2017, dou_chapter_2020, ateeq_ensemble-classifiers-assisted_2018, electronics10182208, li_detecting_2021, liu_cerebral_2020, lu_detection_2017, Myung2021NovelPerformance, stanley_automated_2022, sundaresan_automated_2022,sa-ngiem_cerebral_2019,afzal_transfer_2022, tajudin_microbleeds_2017}.
Datasets used in the first category of researches focused on AD \cite{kirsch_serial_2009, mayeux_washington_nodate, ellis_australian_2009, hofman_rotterdam_2007, noauthor_alzheimers_nodate, noauthor_adni_nodate}, SMART \cite{simons_second_1999}, TBI \cite{pacurar_database_2016}, stroke \cite{pacurar_database_2016, rothwell_change_2004}, ICH \cite{odonnell_apolipoprotein_2000, sprigg_tranexamic_2018, dineen_does_2018}, gliomas \cite{morrison_user-guided_2018_github}, hemodialysis cases \cite{pacurar_database_2016}, CAA \cite{chen_progression_2006}, atherosclerosis \cite{heckbert_yield_2018}, or CMBs \cite{noauthor_uk_nodate}.

Clinicians rated the CMBs present in the images from these datasets according to the MARS scale, the BOMBS scale, or an unspecified standard. Details related to the number of patients, image acquisition parameters, types of sequences, strength of magnetic field and data availability are given in Table \ref{datasets_comparison_table}. The abbreviations used in the table stand for: RES -- resolution, TR -- repetition time, TE -- echo time, FA -- flip angle, BW -- bandwidth, IMS -- image matrix size, ST -- slice thickness, FOV -- field of view, u - unknown dimension.

Datasets that were not included in the table due to insufficient information are \cite{tajudin_microbleeds_2017, k_standvoss_cerebral_2018, doke_using_2020, lu_diagnosis_2020, bao_voxelwise_2018, tao_voxelwise_2018, lu_detection_2017, gunter_p4-232_2018}. They contained only information about, for example, the number of patients or the type of sequence.

\section{Methodology}
\label{sec:methodology}
A comprehensive analysis of the past works regarding cerebral microbleeds detection has led to the proposition of a generalized pipeline of such a system. The majority of works can be divided into three stages: \textit{Pre-processing}, \textit{CMB Candidates Detection} and \textit{CMB Candidates Verification}.
Therefore, we decided to describe the methodology having regard to such division. The overall idea is presented in Fig. \ref{fig:pipeline}.
All the methods and algorithms available within each stage are firstly described as single transform, that can be applied. Their further synthesis into a complete approach along with the paper in which it was used are in Table \ref{approaches_comparison_table}.

\subsection{Pre-processing}
\label{sec:pre-processing}
Data pre-processing is an important step in system synthesis.
Proper preparation of data has a significant impact on further system performance. 

It is important to understand that phrase \textit{raw data} does not always mean that data were not pre-processed by MRI software.
From the system's perspective, raw data are those provided by the MRI.
However, there is a variety of MRI device suppliers, who design an operating system of their own, which performs different operations on a particular scan before the image delivery.
Therefore, it is crucial to know how the data has already been processed and what else can be improved to meet system needs.

\begin{figure}
\centering
    \includegraphics[width=0.48\textwidth]{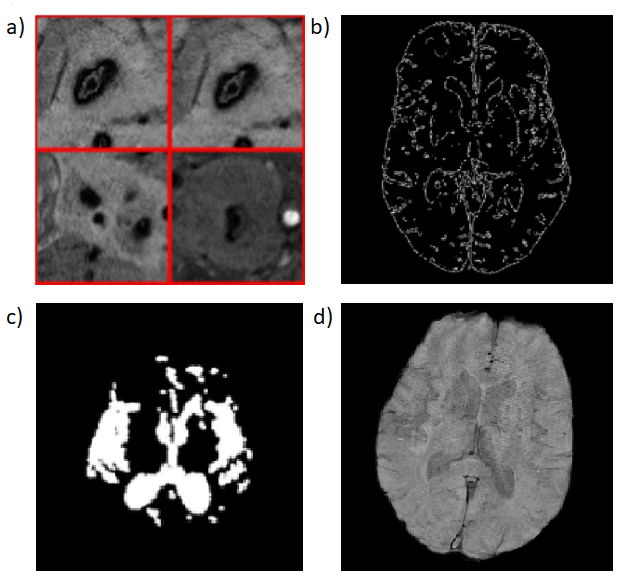}
    \caption{Example of pre-processing operations: a) sliding neighborhood processing \cite{hong_classification_2020}, b) Canny edge detection \cite{chesebro_automated_2021}, c)  CSF mask \cite{Myung2021NovelPerformance} d) brain extraction using BrainSuite software.}
    \label{fig:pre-processing}
\end{figure}

Below the most popular types of operations performed on raw data are presented and few examples illustrated in Fig. \ref{fig:pre-processing}.
Firstly, there are operations for removing artifacts and unnecessary information.

\textbf{Bias field correction} is the operation that reduces negative influence of the bias field, which is an undesired artifact in most MRI images, especially old ones. It can also be called \textbf{intensity inhomogeneity correction}.
The most commonly known techniques include N3 Bias Correction \cite{sled_nonparametric_1998} and its successor N4ITK/N4 \cite{tustison_n4itk_2010}, FSL FAST \cite{zhang_segmentation_2001} or reconstruction Syngo MR B17 provided by the manufacturer.
Nevertheless, there are also other methods for bias field correction \cite{song_review_2017, LUPO2009480}. This operation was applied by \cite{ghafaryasl_computer_2012,fazlollahi_efficient_2014,fazlollahi_computer-aided_2015,ourselin_cerebral_2015,t_l__a_van_den_heuvel_computer_2015,zhang_sparse_2016,van_den_heuvel_automated_2016,liu_cerebral_2019,lu_diagnosis_2020,rashid_deepmir_2021,momeni_synthetic_2021, sundaresan_automated_2022}.
\textbf{Skull stripping} also known as \textbf{brain extraction} is an operation of removing skull and background from the image, leaving only the brain.
There are plenty of algorithms for performing this task: Brain Extraction Tool (BET) \cite{smith_fast_2002}, BrainSuite \cite{shattuck_brainsuite_2002}, and others \cite{carass_simple_2011, skull-strip2}. Brain extraction was applied in \cite{kuijf_detecting_2011,barnes_semiautomated_2011,ghafaryasl_computer_2012,bian_computer-aided_2013,fazlollahi_efficient_2014,ourselin_cerebral_2015,t_l__a_van_den_heuvel_computer_2015,van_den_heuvel_automated_2016,ateeq_ensemble-classifiers-assisted_2018,morrison_user-guided_2018,chen_toward_2019,al-masni_automated_2020,al-masni_two_2020,chesebro_automated_2021,Myung2021NovelPerformance,afzal_transfer_2022,sundaresan_automated_2022}.

\textbf{Normalization} is a typical operation of rescaling the pixel values into range (0,1) or (-1,1).
This enables bias reduction in the next stages of system creation. It was applied by \cite{kuijf_detecting_2011,seghier_microbleed_2011,barnes_semiautomated_2011,kuijf_efficient_2012,kuijf_semi-automated_2013,bian_computer-aided_2013,fazlollahi_automatic_2013,fazlollahi_efficient_2014,fazlollahi_computer-aided_2015,chen_automatic_2015,qi_dou_automatic_2015,t_l__a_van_den_heuvel_computer_2015,liu_cerebral_2019,rashid_deepmir_2021,electronics10182208,stanley_automated_2022}.

\textbf{Standardization} is an equally common operation as normalization and involves subtraction of mean value of pixels and division by the standard deviation of them. It was claimed to be used in \cite{kuijf_detecting_2011,kuijf_efficient_2012,kuijf_semi-automated_2013,electronics10182208}.

\textbf{Mask generation} is a wide term as different types of masks might be generated. The most common is binary mask that might be generated using Statistical Parametric Mapping Toolbox \cite{mathworks_spm_2020} or morphological operations \cite{soille_morphological_2004, seghier_lesion_2008}. Further, there are typically neurological masks such as cerebrovascular fluid (CSF) mask, gray-white matter (GWM) mask and white-matter (WM) mask. Masks were generated in \cite{kuijf_detecting_2011,seghier_microbleed_2011,kuijf_efficient_2012,kuijf_semi-automated_2013,bian_computer-aided_2013,fazlollahi_automatic_2013,fazlollahi_efficient_2014,fazlollahi_computer-aided_2015,t_l__a_van_den_heuvel_computer_2015,van_den_heuvel_automated_2016,gunter_p4-232_2018,chen_toward_2019,sa-ngiem_cerebral_2019,hong_classification_2020,liu_cerebral_2020,chesebro_automated_2021,stanley_automated_2022}.

\textbf{Further image generation} involves using images provided by the MRI device to make a new image consisting more information.
For instance, a SWI sequence is generated from the Magnitude and Phase sequences.
These days, it is the standard sequence generated by the scanner.
Further, the SWI data might be processed using \cite{li_quantitative_2014} for phase enhancement, like in \cite{ourselin_cerebral_2015}
Similarly, T2*-weighted images are nowadays provided by the MRI scanner, but in the past they had to be obtained from PD-weighted images using, for example, Elastix Tool \cite{klein_elastix_2010}. It was performed for example in \cite{ghafaryasl_computer_2012,liu_cerebral_2019, chen_toward_2019}.
A QSM image can be generated using Morphology Enabled Dipol Inversion (MEDI) \cite{liu_morphology_2012}, like in \cite{rashid_deepmir_2021}.

\textbf{Slice merging} can also be considered as a new image creation, which involves concatenation of adjacent slices to provide 3D information.
Generally, MRI images are one-channel.
It enables putting three adjacent slices into one image using RGB channels.
The concatenation of different sequences of corresponding slices might be done as well.
However, in this case it may be necessary to align the slices with each other, if there were different parameters of acquisition.
This kind of operation was performed in \cite{dou_automatic_2016,al-masni_automated_2020,al-masni_two_2020,electronics10182208}.

Useful software to perform these operations is Neuroimaging Core \cite{dianne_patterson_neuroimaging_2019} involving Advanced Normalization Tools (ANT), FMRIB Software Library (FSL) \cite{ashburner_unified_2005, jenkinson_global_2001, jenkinson_improved_2002, jenkinson2012fsl} and Statistical Parametric Mapping (SPM). The last software is also implemented in \cite{mathworks_spm_2020} based on \cite{penny_statistical_2006}.

There are also some typical transforms performed in standard image pre-processing.
It is noteworthy that medical data are highly sensitive to any transformation, after which significant information can be accidentally lost.

\textbf{DICOM to JPG conversion} is an excellent example of lossy data conversion technique, which might influence further processing stages. It was done by \cite{li_detecting_2021}.
Although DICOM or NIFTY formats might be considered not developer--friendly, working on original image matrices should be a standard.

\textbf{Resize} is a common operation of changing image size.
It is usually performed to obtain equal sizes of all images, or to enlarge images so that the objects were more visible. It can also be forced by requirements of a method used in the \textit{CMB Candidates Detection} stage. The images were resized in \cite{barnes_semiautomated_2011,liu_cerebral_2019, chesebro_automated_2021,electronics10182208}.

\textbf{Padding} is performing an artificial size change by addition of a black frame to obtain a desired image size without applying resize. It was utilized by \cite{rashid_deepmir_2021,electronics10182208,stanley_automated_2022}.

\textbf{Image cut} is a common operation performed to simplify the detection task.
It involves image partitioning into smaller parts and then feeding them into the classifier.
It might be performed using the sliding neighborhood processing (SNP) technique to produce smaller fragments of the original image. A lot of works utilized this method: \cite{zhang_sparse_2016,lu_detection_2017,wang_cerebral_2017,zhang_seven-layer_2018,zhang_voxelwise_2018,bao_voxelwise_2018,tao_voxelwise_2018,wang_cerebral_2019,hong_detecting_2019,hong_improvement_2019,hong_classification_2020,doke_using_2020,lu_diagnosis_2020,lu_acerebral_2021,lu_cerebral_2021}.

\textbf{Rotation} is a simple operation of changing image orientation. However, it can be a loss operation, and therefore a rotation with original intensity should be considered, like in \cite{sundaresan_automated_2022}, for instance by using - fslreorient2std tool \cite{jenkinson2012fsl}.

\textbf{Inversion} is the operation which consists of swapping intensity values in relation to the center of the intensity interval and it was performed by \cite{fazlollahi_computer-aided_2015}.

Finally, there is \textbf{data augmentation}, which is not always considered as a pre-processing technique, but rather a regularization one.
However, it is sometimes performed at this stage and consists of image transformations, therefore it is placed in this section.
It enhances a dataset, especially in case of small amount of data by creating new, slightly modified, artificial images. There is a wide range of transformations, including those described above, along with blur, crop, etc. \cite{mikolajczyk_data_2018,albumentations,pytorch}. Augmentation was used in \cite{k_standvoss_cerebral_2018,gunter_p4-232_2018,doke_using_2020,rashid_deepmir_2021,Myung2021NovelPerformance,li_detecting_2021,electronics10182208,momeni_synthetic_2021,afzal_transfer_2022}.

\subsection{Algorithms for CMB candidates detection}
\label{sec:cmb_detection}
Over the years, a wide range of algorithms were used to detect cerebral microbleeds, starting from the simplest methods based on traditional image transformations, up to complicated deep learning models.

\subsubsection{Classical methods}

In early works regarding CMBs detection the candidates were extracted using predetermined features such as: \textbf{intensity threshold} and \textbf{area size} \cite{barnes_semiautomated_2011,ghafaryasl_computer_2012,chen_automatic_2015,qi_dou_automatic_2015,ateeq_ensemble-classifiers-assisted_2018}.
In the SWI sequence CMBs occur as low-intensity spheres, therefore applying a proper intensity threshold allows for \textbf{binary mask} generation.
Sometimes the authors also applied morphological operations such as \textbf{filtering}, \textbf{hole filling}, etc. \cite{seghier_microbleed_2011,ateeq_ensemble-classifiers-assisted_2018,afzal_transfer_2022}.
However, this kind of operations were used at all the stages described in this paper, as they were also useful for \textit{CMB candidates verification}.
Next, the detecting procedures evolved to include more complicated voxel features, such as: \textbf{eigenvalues} in \cite{sundaresan_automated_2022} -- scalars associated with the given linear transformation, \textbf{line detection} in \cite{fazlollahi_automatic_2013} -- defining the line where edge points are located, \textbf{Gaussian filter} in \cite{sundaresan_automated_2022} and \textbf{Laplacian of Gaussian} operator in\cite{fazlollahi_efficient_2014,fazlollahi_computer-aided_2015,sundaresan_automated_2022}, which highlight the rapid change of the image intensity, \textbf{Hough transform} in \cite{chesebro_automated_2021} that enables shape detection by finding objects - local maxima, \textbf{Canny filter} in \cite{chesebro_automated_2021} which enables edge detection \textbf{watershed transform} in \cite{tajudin_microbleeds_2017} - transforming images to grayscale topographic like map, and distinguishing objects on the basis of its intensity value or \textbf{Frangi filters} in \cite{sundaresan_automated_2022} - a dedicated filter enabling vessel distinction, or 3D gradient co-occurrence matrix (3D GCM) in \cite{stanley_automated_2022}, which indicates the differences between intensity of two adjacent pixels.

Simultaneously, the researchers began to use the Radial Symmetry Transform (RST), and its successor Fast Radial Symmetry Transform (FRST) \cite{loy_fast_2003}.
This algorithm deserves special attention since it is successfully used to this day \cite{kuijf_detecting_2011,kuijf_efficient_2012,kuijf_semi-automated_2013,bian_computer-aided_2013,morrison_user-guided_2018,liu_cerebral_2019,chen_toward_2019,sundaresan_automated_2022}.
In this transform, a gradient of the image is computed, then the orientation and magnitude of each pixel is established.
Next, using the above values, points of interest can be selected according to the given formula.
This algorithm was later developed so that it could be used in 3D space.
However, despite its common use, when applied to candidates detection, FRST generates a lot of false positives, which forces introducing the third stage to the whole detection procedure.

Another algorithm used in the \textit{CMB Candidates Detection} stage by \cite{ourselin_cerebral_2015} was the \textbf{region growing} that inspects the homogeneity of the considered pixel - or voxel in case of 3D \cite{revol-muller_automated_2002}. 

\subsubsection{Neural networks-based methods}

Then, with the development of neural networks (NN), algorithms based on these networks gained more attention.
Generally, two approaches for neural networks usage may be distinguished: custom or general purpose pretrained neural network
In the first case, in the domain of CMBs detection various approaches were used, such as: simple \textbf{artificial neural networks} (ANN) \cite{wang_artificial_2003} used in \cite{zhang_seven-layer_2018,momeni_synthetic_2021}, which is basically a sum of inputs multiplied by weights assigned in the training process, \textbf{back-propagation neural networks} (BPNN) \cite{buscema_back_1998} utilized in \cite{tao_voxelwise_2018} that are ANNs extended with the information about the error, \textbf{sparse auto-encoder} (SAE) \cite{lee_efficient_nodate} used in \cite{zhang_sparse_2016,zhang_voxelwise_2018} that is a neural network consisted of encoder and decoder with the additional sparsity penalty algorithm.

The \textbf{Random Forest algorithm} \cite{breiman_random_2001} was used as well in \cite{t_l__a_van_den_heuvel_computer_2015,van_den_heuvel_automated_2016}. It is a black-box algorithm that consists of an ensemble of classifiers that predict an output value based on a part of a dataset and then these predictions are averaged into one.

Finally, there are \textbf{convolutional neural networks} (CNN) \cite{oshea_introduction_2015}, which are the most popular solution \cite{lu_detection_2017,wang_cerebral_2017,gunter_p4-232_2018,hong_improvement_2019,hong_classification_2020,doke_using_2020,lu_acerebral_2021,lu_cerebral_2021,stanley_automated_2022}.

Basically, CNN consists of a number of feed-forward convolutional layers, where the features are extracted by performing a convolution with predefined filters on every image and then further modified during training.
Each convolution layer is followed by a non-linear activation function.
Consecutive convolution layers are interspersed by pooling layers that extract the most important features.
Then, mostly, there is a fully connected layer or other classifier that assigns a predicted class based on the previously extracted features.
An interesting approach is replacement of a fully-connected layer by Extreme Learning Machine \cite{huang_extreme_2006}, which is much more efficient \cite{lu_cerebral_2021}.

In the second approach, one takes advantage of a deep neural network architecture that has already been trained on a vast dataset -- transfer-learning -- often very different from terminal one and just adjusts it to the considered problem.
These networks usually consist of millions of parameters and are hard to train on the CPU due to hardware limitations.
Additionally, it is a good method for dealing with small dataset problem.
The transfer-learning idea is to use a pre-trained network that has already learned some image features and fine-tune it on the particular dataset.
Several networks were used for this purpose, including: AlexNet \cite{krizhevsky_imagenet_2012} in \cite{sa-ngiem_cerebral_2019}, ResNet50 \cite{resnet50} in \cite{hong_detecting_2019}, Faster-RCNN \cite{ren_faster_2015} in \cite{electronics10182208}, VGG \cite{simonyan_very_2015} in \cite{lu_diagnosis_2020}, U-Net \cite{ronneberger_u-net_2015} in \cite{rashid_deepmir_2021}, YOLOv2 \cite{Redmon_2017_CVPR} in \cite{al-masni_automated_2020,al-masni_two_2020}, DenseNet 201 \cite{huang_densely_2018} in \cite{wang_cerebral_2019} or SSD \cite{leibe_ssd_2016} in \cite{li_detecting_2021} with the modification of feature enhancement.
Sometimes, especially in case of SNP algorithm usage the detection task was substituted by classification of small fragments of image using either CNN or ResNet50 for instance in \cite{hong_detecting_2019,hong_improvement_2019}.
Considering the main aim of this paper, the description of each network is omitted, as they are explained in detail in the mentioned papers.
Nevertheless, the reader is strongly encouraged to get familiar with these architectures.

Relatively new and still not fully explored architectures are \textbf{3D convolutional neural networks} (3D CNN). The idea is the same as in 2D CNNs, but instead of performing convolution on 2D matrices, it is performed on 3D patches. They were applied in \cite{dou_automatic_2016, k_standvoss_cerebral_2018}.

\subsection{Algorithms for CMB candidates verification}
\label{sec:cmb_verification}
Due to the nature of the considered problem, most of the presented approaches involved the \textit{CMB Candidates Verification} stage.
In spite of this, some solutions still have not managed to acquire satisfying quality.

In some cases, the process of false positive candidates elimination was performed manually by a radiologist \cite{kuijf_detecting_2011,barnes_semiautomated_2011,kuijf_efficient_2012,kuijf_semi-automated_2013,morrison_user-guided_2018,chen_toward_2019,chesebro_automated_2021}.
Although this kind of approach significantly reduced the time needed for one scan rating, it is a semi-automated one.

A large part of the research involved at this stage establishing a batch of predefined features of CMB : from simple ones as intensity or size, to very complicated parameters of a single voxel, calculated in 2D or 3D spaces.
There were also other methods for defining the feature vectors, for instance 2D CNN in \cite{chen_automatic_2015, afzal_transfer_2022}, 3D ISA network \cite{Comon95supervisedclassification:} in \cite{qi_dou_automatic_2015}, 3D Radon Transform \cite{averbuch_3d_2003} in \cite{fazlollahi_efficient_2014,fazlollahi_computer-aided_2015} or feed-forward feature selection (FFFS) \cite{luo_methods_2011} in \cite{ghafaryasl_computer_2012}.

In some cases, thresholds of geometric features were set, and on this ground the classification was performed \cite{bian_computer-aided_2013,ourselin_cerebral_2015,morrison_user-guided_2018_github, chesebro_automated_2021,sundaresan_automated_2022}.

In others, these features together with the previously prepared fragments of images were passed to the classifier.
A lot of classifiers have already been tested: Supported Vector Machine (SVM) \cite{chang_libsvm_2011} in \cite{barnes_semiautomated_2011, chen_automatic_2015, qi_dou_automatic_2015, ateeq_ensemble-classifiers-assisted_2018}, linear criterion classifier (LDC) \cite{linear_classifier}, quadratic discriminant classifier (QDC) \cite{tharwat_linear_2016}, Parzen classifier \cite{gelsema_classifier_1988} in \cite{ghafaryasl_computer_2012} and Random Forrest Classifier (RFC) \cite{breiman_random_2001} in \cite{fazlollahi_efficient_2014,fazlollahi_computer-aided_2015}.

A common approach at this point was also using a previously generated CSF mask to distinguish a real CMB from vessels, and a WM mask to include the information about the location of potential microbleed \cite{bian_computer-aided_2013, chesebro_automated_2021, Myung2021NovelPerformance}.

Another approach utilized the advantage of a 3D CNN. It was usually performed for 3D information inclusion, resulting with FP reduction, after the 2D algorithm used in the \textit{Candidates detection stage} \cite{singh_3d_2020} \cite{dou_automatic_2016, liu_cerebral_2019, chen_toward_2019, al-masni_automated_2020, al-masni_two_2020}.

Some works present also usage of region growing algorithm for CMB verfication \cite{bian_computer-aided_2013, van_den_heuvel_automated_2016, morrison_user-guided_2018_github}.

There was also an algorithm investigating the overlap between predictions from adjacent slices \cite{electronics10182208}. It not only enabled removal of false positive predictions that were in fact a ground truth, although labeled in the adjacent slice, but also helped finding a real CMB that was detected in the adjacent slice in spite of the previous false negative prediction.

\subsection{System output and evaluation}
\label{sec:evaluation}
To comprehensively validate the quality and robustness of the system, one should take advantage of a number of commonly accepted metrics that provide complementary insight into various aspects of system performance. 

A common oversight is to not include metrics that are complementary and provide a view of the system as a whole, not just a part of it.
For instance, the sensitivity metric is useless alone, as it can be artificially inflated. It is necessary to provide the precision or F1 score value to properly interpret the sensitivity.
In addition, the lack of a uniform way of result evaluation makes it impossible to compare approaches and effectively assess their usefulness.

The evaluation should be performed on a separate dataset or at least separate subjects, using, for instance, cross-validation to avoid randomness.

There are different metrics regarding the type of solved problem.
For classification evaluation, the most popular metrics are accuracy (\ref{accuracy}), precision (\ref{precision}), sensitivity/recall (\ref{sensitivity}), and F1 score (\ref{F1}) that combines precision and sensitivity.

In the case of detection and segmentation, more detailed metrics are required as not only a proper class is important, but also the overlapped area of ground truth label and prediction.
In that case, the average precision (\ref{AP}) metric is used, and it is calculated for different values of IoU (\ref{IOU}).

The CMBs detection task is known to produce large number of false positive predictions. Therefore two additional metrics were provided particularly for this problem, namely it is FPavg (\ref{FP}) and FPcmb (\ref{FPcmb}).

The mentioned metrics are calculated as follows: 

\begin{equation} \label{accuracy}
\textrm{accuracy} = \frac{TP+TN}{TP+TN+FP+FN}
\end{equation}
\begin{equation} \label{sensitivity}
\textrm{sensitivity} = \frac{TP}{TP+FN}
\end{equation}
\begin{equation} \label{specificity}
\textrm{specificity} = \frac{TN}{TN+FP}
\end{equation}
\begin{equation} \label{precision}
\textrm{precision} = \frac{TP}{TP+FP}
\end{equation}
\begin{equation} \label{F1}
\textrm{F1 score} = 2\times\frac{\textrm{sensitivity}\times
\textrm{precision}}{\textrm{sensitivity}+\textrm{precision}}
\end{equation}
\begin{equation} \label{IOU}
\textrm{IoU} = \frac{Overlap area}{Union area}
\end{equation}
\begin{equation} \label{AP}
\textrm{AP} = \int_{0}^{1} p(r) \,dr
\end{equation}
\begin{equation} \label{FP}
\textrm{FPavg} = \frac{FP}{n}
\end{equation}
\begin{equation} \label{FPcmb}
\textrm{FPcmb} = \frac{FP}{m}
\end{equation}
where:
\begin{itemize}
\item $TP$---true positive -- the number of actual CMBs detected;
\item $FP$---false positive -- the number of predicted CMBs that were not marked as CMB in ground truth;
\item $FN$---false negative -- the number of actual CMBs not detected;
\item $IoU$---intersection over union;
\item $r$---recall (sensitivity);
\item $p(r)$---precision as function of recall;
\item $n$--- number of subjects (patients) in the test set;
\item $m$--- number of CMBs in the test set.
\end{itemize}

Accuracy (ACC) (\ref{accuracy}) shows how the system deals with the classification in general.
A high score means that almost all labels have been properly assigned.

Sensitivity/recall (\ref{sensitivity}), also known as true positive rate (TPR), shows how the system deals with the ground truth detection or classification.
A high score means that almost all ground-true samples have been determined.

Specificity, also known as true negative rate (TNR) (\ref{specificity}), discloses the system ability to recognize the negative class.

Precision (\ref{precision}), or positive predictive value (PPV), informs whether the prediction matches ground truth. A high score means that the system generates a small number of false positives.

F1 score (\ref{F1}) helps to check whether there is a balance between sensitivity and precision.

IoU (\ref{IOU}) stands for Intersection over Union and shows the common area between prediction and ground truth.
It is actually a special case of geometrically oriented Jaccard Index \cite{real_probabilistic_1996}.
The average precision (\ref{AP}) AP@0.5 represents the area under the precision-recall curve with IoU of 0.5 and it is used in detection and segmentation.
There is also an AUC - area under curve - metric.
In case of classification it refers to the ROC curve - sensitivity as a function of 1-specificity.

FPavg (\ref{FP}) shows the average number of false positive predictions per subject, while FPcmb (\ref{FPcmb}) is the number of false positive predictions per one ground truth sample. For example, when we have one subject with 5 ground truth CMBs and 1 false positive prediction. The FPavg will equal 1 and FPcmb will equal 0.2.

\begin{landscape}
\begin{savenotes}
\begin{longtable}{c|c c c|c c c c c c c c c}
    \caption{Comparison of existing  approaches \protect\footnote{Data marked with * were not provided in original paper. Instead, they were calculated either by us or the Authors of other papers listed in Table \ref{approaches_comparison_table}, based on data provided in the original paper.} The most promising ones are marked with bold.} \label{approaches_comparison_table}\\

    \textbf{Reference} & \textbf{Pre-processing} & \textbf{First stage} & \textbf{Second stage} & \textbf{TPR} & \textbf{PPV} & \textbf{F1} & \textbf{FPavg} & \textbf{FP/CMB} & \textbf{TNR} & \textbf{ACC}\\
    \hline
    Kuijf et al., & SPM8, BET,& 3D RST & manual & - & - & - & 5* & - & - & -\\
    2011, \cite{kuijf_detecting_2011} & normalization, & & inspection & & & & & & &\\
    & standardization & & & & & & & & &\\
    \hline
    Seghier et al.,  & SPM8, & CSF, GWM, CMBs, & morphological & \multicolumn{7}{c}{Authors did not provide any metric, only the table}\\
    2011, \cite{seghier_microbleed_2011} & normalization & skull scalp, & operations & \multicolumn{7}{c}{of results for each case.}\\
    & & background img & (2 iterations) & & & & & & &\\
    \hline
    Barnes et al., & brain extraction, & intensity histogram & SVM, manual & 81.7 & - & - & 107.5* & 5.4* & 100 & -\\
    2011, \cite{barnes_semiautomated_2011} & resize, & threshold & review & & & & & & &\\
    & normalization & & & & & & & & &\\
    \hline
    Ghafaryasl et al., & N3, Elastix, & intensity and & FFFS \textrightarrow  LDC, & 90.9 & - & - & 4.1 & 1.8* & - & -\\
    2012, \cite{ghafaryasl_computer_2012} & BET & area threshold & QDC, SVC, & & & & & & &\\
    & & & Parzen & & & & & & &\\
    \hline
    Kuijf et al., & SPM8,& 3D RST & manual & 71.2 & - & - & 17.17 & 4.68* & - & -\\
    2012, \cite{kuijf_efficient_2012} & normalization, & & inspection & & & & & & &\\
    & standardization & & & & & & & & &\\
    \hline
    Kuijf et al., & SPM8,& 3D RST & manual & 87 & - & - & 45 & - & - & -\\
    2013, \cite{kuijf_semi-automated_2013} & normalization, & & inspection & & & & & & &\\
    & standardization & & & & & & & & &\\
    \hline
    Bian et al., & BET, ARC, mIP, & FRST & vessel mask & 86.5 & - & - & 44.9 & 1.5* & - & -\\
    2013, \cite{bian_computer-aided_2013} & normalization & & screening, & & & & & & &\\
    & & & 3D region & & & & & & &\\
    & & & growing, & & & & & & &\\
    &  & & geometric & & & & & & &\\
    &  & & features & & & & & & &\\
    \hline
    Fazlollahi et al., & CSF, & multi-scale 1D & center & 100 & - & - & 158.93* & - & 99.9 & -\\
    2013, \cite{fazlollahi_automatic_2013} & invertion, & line detection & detection \textrightarrow & & & & & &\\
    & normalization, & & Hessian & & & & & & &\\
    & Gaussian blur & & matrix & & & & & & &\\
    \hline
    Fazlollahi et al., & N4, CSF, & multi-scale & 3D Rodon & 92.04 & - & - & 16.84 & 6.7* & - & -\\
    2014, \cite{fazlollahi_efficient_2014} & skull-stripping, & Laplacian & Transform \textrightarrow & & & & & &\\
    & normalization, & of Gaussian & Hessian & & & & & & &\\
    & equalization, & & matrix, & & & & & & &\\
    & anisotropic & & RFC & & & & & & &\\
    & diffusion & & & & & & & & &\\
    \hline
    Fazlollahi et al., & N4, CSF, & Laplacian & 3D Rodon & 87 & - & - & 27.1 & - & - & -\\
    2015, \cite{fazlollahi_computer-aided_2015} & inversion, & of Gaussian & Transform \textrightarrow & & & & & &\\
    & normalization, & & Hessian & & & & & & &\\
    & equalization, & & matrix, & & & & & & &\\
    & anisotropic & & RFC & & & & & & &\\
    & diffusion & & & & & & & & &\\
    \hline
    Roy et al., & N4, & 3D region & RST, & 85.7 & - & - & - & - & 99.5 & -\\
    2015, \cite{ourselin_cerebral_2015} & skull stripping, & growing & WM mask, & & & & & & &\\
    & phase & & geometric & & & & & & &\\
    & enhancement & & features & & & & & & &\\
    \hline
    Chen et al., & normalization & intensity & CNN, & 89.13 & 56.16 & 68.91 & 6.4 & - & - & -\\
    2015, \cite{chen_automatic_2015} & & threshold & 3D concatenation, & & & & & & &\\
    & & & SVM & & & & & & &\\
    \hline
    Dou et al., & normalization & intensity & ISA & 89.44 & - & - & 7.7 & 0.9 & - & -\\
    2015, \cite{qi_dou_automatic_2015} & & threshold & SVM & & & & & & &\\
    \hline
    van den & FSL FLIRT, & voxel based & - & 90 & - & - & - & 1.3 & - & -\\
    Heuvel et al., & FSL FAST, & features \textrightarrow & & & & & & & &\\
    2015, \cite{t_l__a_van_den_heuvel_computer_2015} & N3, SPM12b, & RFC & & & & & & & &\\
    & normalization & & & & & & & & &\\
    \hline
    Dou et al., & slices & hierarchical & 3D CNN & 93.16 & 44.31 & 60.06 & 2.74 & - & -\\
    2016, \cite{dou_automatic_2016} & merging & 3D CNN & & & & & & & &\\
    \hline
    Zhang et al., & reconstruction & SAE & - & 93.20 & - & - & - & - & 93.25 & 93.22\\
    2016, \cite{zhang_sparse_2016} & Syngo MR B17, & & & & & & & & &\\
    & SNP & & & & & & & & &\\
    \hline
    van den & FSL FLIRT, & voxel based & object & 93 & - & - & 25.9 & 0.29 & - & -\\
    Heuvel et al., & FSL FAST, & features \textrightarrow & classifier, & & & & & & &\\
    2016, \cite{van_den_heuvel_automated_2016} & N3, SPM12b & RFC & growing-based & & & & & & &\\
    & & & algorithm & & & & & & &\\
    \hline
    Lu et al., & square & CNN & - & 97.29 & - & - & - & - & 92.23 & 96.05\\
    2017, \cite{lu_detection_2017} & window size & & & & & & & & &\\
    \hline
    Wang et al., & SNP, & CNN+RAP & - & 96.94 & - & - & - & - & 97.18 & 97.18\\
    2017, \cite{wang_cerebral_2017} & discard & & & & & & & & &\\
    & borders, & & & & & & & & &\\
    & cost ratio & & & & & & & & &\\
    \hline
    Tajudin et al., & - & watershed & - & \multicolumn{7}{c}{Authors provided only mean square error MSE = 0.089 and peak}\\
    2017, \cite{tajudin_microbleeds_2017} & & transform, & & \multicolumn{7}{c}{signal to noise ratio PSNR = 34.5221}\\
    & & active contour & & & & & & & &\\
    & & (Chan-Vese) & & & & & & & &\\
    \hline
    Standvoss et al., & augmentation, & 3D CNN, & - & 87 & - & - & 16.75 & 2.5 & - & -\\
    2018, \cite{k_standvoss_cerebral_2018} & selective & connected & & & & & & & &\\
    & sampling & component & & & & & & & &\\
    & & analysis & & & & & & & &\\
    \hline
    Zhang et al., & SNP, & ANN & - & 93.05 & - & - & - & - & 93.06 & 93.06\\
    2018, \cite{zhang_voxelwise_2018} & discard & & & & & & & & &\\
    & borders, & & & & & & & & &\\
    & cost ratio & & & & & & & & &\\
    \hline
    Zhang et al., & SNP, & SAE-DNN & - & 95.13 & - & - & - & - & 93.33 & 94.23\\
    2018, \cite{zhang_seven-layer_2018} & discard & & & & & & & & &\\
    & borders & & & & & & & & &\\
    \hline
    Ateeq et al., & BrainSuite & intensity & SVM, QDA, & 93.7 & - & - & 56 & 5.3 & - & -\\
    2018, \cite{ateeq_ensemble-classifiers-assisted_2018} & & threshold, & ensemble & & & & & & &\\
    & & filtering, & classifier & & & & & & &\\
    & & hole filling & & & & & & & &\\
    \hline
    Morrison et al., & BET & FRST & region & 86.7 & - & - & 44.9 & 1.5* & - & -\\
    2018, \cite{morrison_user-guided_2018} & & & growing, & & & & & & &\\
    & & & geometric & & & & & & &\\
    & & & features, & & & & & & &\\
    & & & manual & & & & & & &\\
    & & & validation & & & & & & &\\
    \hline
    Bao et al., & SNP & Bayesian & - & 74.53 & - & - & - & - & 74.51 & 74.52\\
    2018, \cite{bao_voxelwise_2018} & & classifier& & & & & & & &\\
    \hline
    Tao et al., & SNP & GA-BPNN & - & 72.90 & - & - & - & - & 72.89 & 72.90\\
    2018, \cite{tao_voxelwise_2018} & & & & & & & & & &\\
    \hline
    Gunter et al., & intensity & CNN & - & \multicolumn{7}{c}{Authors provided only AUC = 98.5}\\
    2018, \cite{gunter_p4-232_2018} & threshold, & & & & & & & & &\\
    & image cut, & & & & & & & & &\\
    & data & & & & & & & & &\\
    & augmentation & & & & & & & & &\\
    \hline
    Liu et al., & N4, & 3D FRST & 3D CNN & 95.80 & 70.90 & 81.49* & 1.6 & 0.39 & - & -\\
    2019, \cite{liu_cerebral_2019} & SWI generation, & & & & & & & & &\\
    & resize, & & & & & & & & &\\
    & normalization & & & & & & & & &\\
    \hline
    Chen et al., & ARC, BET, & FRST & manual & 94.69 & 71.98 & 81.79 & 11.58 & - & - & - \\
    2019, \cite{chen_toward_2019} & SWI generation, & & validation, & & & & & & &\\
    & negative & & 3D ResNet & & & & & & &\\
    & phase mask & & & & & & & & &\\
    \hline
    \textbf{Wang et al.,} & sliding & Dense-Net 201 & - & \textbf{97.78} & \textbf{97.65} & - & - & - & \textbf{97.64} & \textbf{97.71}\\
    \textbf{2019, \cite{wang_cerebral_2019}} & window & & & & & & & & &\\
    \hline
    Hong et al., & SNP & ResNet50 & - & 95.71 & - & - & - & - & 99.21 & 97.46\\
    2019, \cite{hong_detecting_2019} & & & & & & & & & &\\
    \hline
    Hong et al., & SNP & CNN & - & 98.87 & - & - & - & - & 96.49 & 97.68\\
    2019, \cite{hong_improvement_2019} & & & & & & & & & &\\
    \hline
    Sa-ngiem et al., & intensity & AlexNet, & - & - & - & - & - & - & - & 95.45\\
    2019, \cite{sa-ngiem_cerebral_2019} & enhancement, & brain area & & & & & & & &\\
    & binarization, & extraction & & & & & & & &\\
    & morphological & & & & & & & & &\\
    & operations, & & & & & & & & &\\
    & geometrical & & & & & & & & &\\
    & features & & & & & & & & &\\
    \hline
    Hong et al., & SNP, brain area & CNN & -& 99.74 & - & - & - & - & 96.89 & 98.32\\
    2020, \cite{hong_classification_2020} & enhancement & & & & & & & & &\\
    \hline
    Doke et al., & sliding & CNN & - & 98.97 & 99.66 & - & - & - & 98.14 & 98.54\\
    2020, \cite{doke_using_2020} & window, & & & & & & & & &\\
    & augmentation & & & & & & & & &\\
    \hline
    Liu et al., & binarization. & Fourier & - & 85.2 & 3.2 & - & 69.5 & - & - & -\\
    2020, \cite{liu_cerebral_2020} & noise & descriptor & & & & & & & &\\
    & reduction & & & & & & & & &\\
    \hline
    Lu et al., & reconstruction & VGG-ELM-BAC & - & 93.08 & - & - & - & - & 87.12 & 90.00\\
    2020, \cite{lu_diagnosis_2020} & Syngo MR B17, & & & & & & & & &\\
    & SNP & & & & & & & & &\\
    \hline
    Al-masni et al., & BET, & YOLOv2 & 3D-CNN & 94.32 & 61.94 & 74.78 & 1.42 & - & - & -\\
    2020, \cite{al-masni_two_2020, al-masni_automated_2020} & slices merging & & & & & & & & &\\
    \hline
    Rashid et al., & N4, & U-Net & - & 84 & 59 & - & - & - & - & -\\
    2020, \cite{rashid_deepmir_2021} & QSM generation, & & & & & & & & &\\
    & padding, & & & & & & & & &\\
    & normalization, & & & & & & & & &\\
    & augmentation & & & & & & & & &\\
    \hline
    Chesebro et al., & BET, & Sobel & CSF filtering, & 95.00 & 11.00 & 19.72 & 9.7 & - & - & -\\
    2021, \cite{chesebro_automated_2021} & CSF mask, & filter, & 3D geometric & & & & & & &\\
    & resize & Hough & filtering, & & & & & & &\\
    & & transform & manual & & & & & & &\\
    & & & validation & & & & & & &\\
    \hline
    Myung et al., & BET & YOLO & CSF & 66.90 & 79.75 & 72.76 & 2.15 & - & - & -\\
    2021, \cite{Myung2021NovelPerformance} & augmentation & & filtering & & & & & & &\\
    \hline
    \textbf{Li et al.,} & ANTs, & SSD + FE & - & \textbf{90} & \textbf{79.7} & \textbf{84.54*} & -  & \textbf{0.23} & - & -\\
    \textbf{2021, \cite{li_detecting_2021}} & JPG conversion, & & & & & & & & &\\
    & augmentation & & & & & & & & &\\
    \hline
    \textbf{Ferlin et al.,} & padding, & Faster RCNN & overlap & \textbf{92.62} & \textbf{89.74} & \textbf{90.84} & \textbf{0.24} & - & - & -\\
    \textbf{2021, \cite{electronics10182208}} & resize, & & between & & & & & & &\\
    & normalization, & & slices & & & & & & &\\
    & standardization, & & & & & & & & &\\
    & slices merging, & & & & & & & & &\\
    & annotations & & & & & & & & &\\
    & modification & & & & & & & & &\\
    & augmentation & & & & & & & & &\\
    \hline
    Lu et al., & SNP & CNN+ELM+BA & - & 92.93 & - & - & - & - & 83.35 & 88.56\\
    2021, \cite{lu_cerebral_2021} & & & & & & & & & &\\
    \hline
    \textbf{Lu et al.,} & SNP & CNN+EN & - & \textbf{98.27} & - & - & - & - & \textbf{98.93} & \textbf{98.60}\\
    \textbf{2021, \cite{lu_acerebral_2021}} & & & & & & & & & &\\
    \hline
    Momeni et al., & N4, & ANN & - & 18.6 & 9.2 & - & 3.6 & - & 99.4 & 96.8\\
    2021, \cite{momeni_synthetic_2021} & augmentation, & & & & & & & & &\\
    & synthetic & & & & & & & & &\\
    & CMBs & & & & & & & & &\\
    & generation & & & & & & & & &\\
    \hline
    Afzal et al., & BrainSuite, & K-means & Alex-Net & 97.26 & - & - & - & - & 96.5 & 96.21\\
    2022, \cite{afzal_transfer_2022} & augmentation & clustering, & & & & & & & &\\
    & & geometrical & & & & & & & &\\
    & & features & & & & & & & &\\
    \hline
    \textbf{Stanley et al.,} & resize, & 1D CNN+LSTM & - & \textbf{98.76} & - & \textbf{98.78} & - & - & \textbf{97.21} & \textbf{98.24}\\
    \textbf{2022, \cite{stanley_automated_2022}} & contrast & & & & & & & & &\\
    & stretching, & & & & & & & & &\\
    & normalization, & & & & & & & & &\\
    & Gaussian Filter, & & & & & & & & &\\
    & histogram & & & & & & & & &\\
    & equalization, & & & & & & & & &\\
    & morphological & & & & & & & & &\\
    & operations, & & & & & & & & &\\
    & Sharr gradient,& & & & & & & & &\\
    & 3D GCM & & & & & & & & &\\
    \hline
    Sundersan et al., & fslreorient2std, & Frangi filters, & geometric & 91 & - & - & - & - & 81 & 86\\
    2022, \cite{sundaresan_automated_2022} & FSL FAST, & FRST, intensity& features & & & & & & &\\
    & BET & transformations, & level & & & & & & &\\
    & & eigenvalues, & threshold & & & & & & &\\
    & & Gaussian filter, & & & & & & & &\\
    & & Laplacian & & & & & & & &\\
    & & of Gaussian & & & & & & & &\\
    \hline
\end{longtable}
\end{savenotes}  
\end{landscape}

\subsection{Comparison of existing approaches}

Table \ref{approaches_comparison_table} presents, in chronological order, multiple approaches regarding cerebral microbleed detection that had place in recent years.
It can be observed that firstly, the prevailing solutions were those based on traditional image processing techniques and only later the proposals based on machine learning algorithms have taken the lead.
It can be seen, that they achieved considerably higher performance, both in terms of sensitivity and low false positive generation.
Therefore, it can be assumed that this latter path is more promising regarding practically applicable solutions.
Alternatively, a combination of traditional and ML methods might be considered.

Regarding the pre-processing stage, there are several operations, such as  bias field correction, skull stripping and normalization, that should be done before providing data into the system.
Other transforms may also be used in particular cases, but they are not essential.

An important issue is related with selecting the type of solved problem: whether it should be classification, detection, or segmentation.
A large part of solutions is based on cutting images into smaller fragments and their further classification.
These approaches are reported to have significantly better ability to distinguish CMB from its mimic.
On the other hand, in the case of detection, a lot of false positive predictions are generated, which often forces introducing the second stage, namely false positive reduction or predictions verification, as high false positive generation is the main problem in CMBs detection.
Another challenge that can be overcome by using classification instead of detection is the size of the lesion.
CMBs are small objects, which makes them difficult to find in the original image.

No significant improvement can be seen for 3D CNN over 2D CNN. Probably, a larger training dataset could enable taking benefit from the 3D CNN structure and consequently achieve better results. For now, however, choosing this type of solution is discouraging, due to higher computation cost.

Moreover, it is clearly visible, that the reported research often lack in some metrics. Even if we accept that mentioning all of indicators is not necessary, it is crucial to provide a proper evaluation.

We would like to emphasize that in our opinion, based on the gathered data, it is difficult to state which approach is the best and it is still the area that requires development.
However, we can draw attention to some most promising solutions: in case of classification \cite{stanley_automated_2022, wang_cerebral_2019} and \cite{lu_acerebral_2021} with the best ACC=98.60\%; while in case of detection \cite{li_detecting_2021} and \cite{electronics10182208} with F1=90.84\%.
The mentioned research distinguish also in terms of balanced results - similar values of all metrics, which is an advantage in comparison to for instance \cite{al-masni_automated_2020, al-masni_two_2020, chesebro_automated_2021} that report higher sensitivity, however suffer from a high false positive predictions generation - low precision.

Although \cite{doke_using_2020, hong_classification_2020} also report high accuracy, the datasets used by them were small.
Results reported in \cite{stanley_automated_2022, electronics10182208, li_detecting_2021} were performed on relatively big, but diverse datasets, therefore are hard to compare.
From mentioned proposals only \cite{wang_cerebral_2019} and \cite{lu_acerebral_2021} can be compared, as they used the same, but small dataset.

Nevertheless, the above careful and comprehensive analysis provides an opportunity to formulate some conclusions, outline best practices, and point out the key elements of a reliable automatic cerebral microbleeds detection system.

\section{Discussion}
\label{sec:discussion}
In this section we discuss the most important aspects for automatic CMBs detection system.

As previously mentioned, a base for any automatic system, especially machine learning model, is the data.
Although traditional image processing methods do not require large amount of data for training, just for validation and testing purposes, it is still essential for proper system evaluation.

In Section \ref{sec:data} a range of datasets is listed which were used in all reported approaches.
These datasets differ not only in terms of acquisition parameters, but also by the origin and medical history of patients.
This kind of diversity makes any comparison of newly proposed approaches almost impossible.
Moreover, some datasets have extremely small number of subjects \cite{barnes_semiautomated_2011,kuijf_detecting_2011}.
In that case, the tested subset is not representative enough.
The system trained on such a narrowed dataset will reveal low generalization ability \cite{van_den_heuvel_automated_2016}.
Therefore, a big, diversified dataset is needed and advanced regularization techniques should be applied \cite{nusrat_comparison_2018} to prevent over-fitting.

An interesting approach to overcome the data shortage problem was proposed by Momeni et al. \cite{momeni_synthetic_2021}.
It consisted of synthetic microbleeds generation based on previously extracted CMB features.
Another way to produce huge amounts of synthesized data are Generative Adversarial Networks (GANs) \cite{creswell_generative_2018}.
They are able to create new images based on the features automatically extracted from the existing, real dataset.
However, again, it is a method that requires relatively large dataset at the beginning.
Despite the risk of biased data generation, both approaches seem promising for extending datasets, next to other augmentation methods.

A good practice regarding a general system evaluation is using a completely unrelated dataset for testing to ensure that the obtained results are impartial like in \cite{electronics10182208,momeni_synthetic_2021}.
The term \textit{unrelated dataset} means the data acquired from a different MRI machine, from subjects of different origin and medical history, and ranked by another rater. Examinations performed on various MRI machines may differ in parameters. It is important to synthesize system resistant to features that do not have a direct impact on prediction. Usage of various datasets ensures insight into the model's generalization ability.
This, however, is an ideal situation, not always achievable in practice.

There are also methods such as k-fold validation that may enable better evaluation within the same dataset.
It is recommended especially during system development as depending on chosen training, validation and testing sets, the obtained results may differ \cite{al-masni_automated_2020, al-masni_two_2020,electronics10182208}.
Another idea is preparation of a system nameplate with detailed description of system properties and target data type and it should point out operating conditions of the system and its limitations.

The unavailability of the used datasets is another limitation in terms of approach comparison.
Although there are many legal restrictions regarding medical data sharing, establishing a benchmark dataset would significantly trigger the development in this domain \cite{leming_construction_2022}, similarly as it was in case of brain glioma segmentation \cite{baid_rsna-asnr-miccai_2021} or determining skeletal age \cite{halabi_rsna_2019,siegel_what_2019}.

Another aspect is the pre-processing stage of system synthesis.
Subjecting images to any transformations should be well--thought and justified.
Considering the risk of valuable data loss, precautions should be taken to prevent that.

A common method of pre-processing is bias field correction as it enables restoring some important information.
Using dedicated tools for skull stripping seems to be a better approach than simply removing part of the image just as in \cite{zhang_voxelwise_2018, zhang_seven-layer_2018, wang_cerebral_2017}.
Although unintentionally, the image passed to the system may still be deformed or partial.
Any operations that modify the size of the image, should be performed without content loss just as in \cite{sundaresan_automated_2022}.

When it comes to system designing, there are several issues that should be considered.
Firstly, the MRI data is given in the three-dimensional space.
Regardless the used algorithms, at some stage it is inevitable to use the information from the third dimension.
Especially, when detecting cerebral microbleeds is concerned, that kind of data is very important, as it enables distinguishing the CMBs from their most common mimics - vessels \cite{bian_computer-aided_2013, fazlollahi_efficient_2014, chen_automatic_2015,dou_automatic_2016, liu_cerebral_2019,chen_toward_2019, al-masni_automated_2020, chesebro_automated_2021, electronics10182208}.
While vessels can be distinguished based on 3D information, the other CMB mimic - calcification looks similar also in the 3D space considering its shape.
In such case, other MRI sequences, except SWI, may be helpful \cite{al-masni_automated_2020}.

In the majority of reported research, the solution process is divided into three stages: \textit{Pre-processing}, \textit{CMB Candidates Detection}, and \textit{CMB Candidates Verification} (Table \ref{approaches_comparison_table}).
This approach is caused by similarities between CMBs and their mimics, with consequent high production of false positive candidates.
All this compels the use of \textit{CMB Candidates verification} stage to eliminate FP candidates.
It may extend the computation time, but it is necessary to obtain satisfying results.
Still, keeping the balance between accuracy and efficiency is important, particularly when real-time usage is concerned.

The next, worth considering issue is the nature of cerebral microbleeds.
They are small hemorrhages, which are sometimes difficult to notice even for an experienced radiologist.
Therefore, the system to be designed should be sensitive to small objects.
For this purpose, automatic systems may turn out even better, as they are able to consider the information that is not visible for human eye.
It is also important to note that more accurate and sensitive MRI machines with properly adjusted parameters increase the chance for finding all microbleeds \cite{nandigam_mr_2009}.

Another problem is related to the possibility of missing some CMBs by an experienced rater.
In any research regarding detection a ground truth has to be established.
However, this is extremely difficult, as the rater agreement  may be at a relatively low level, for instance - $\kappa=0.68$ \cite{seghier_microbleed_2011}.
To reduce this problem, preliminary rating should be performed by as many raters as possible.
Additionally, verification of system results may be helpful, as some missing CMBs may be detected by the system and should not be treated as false positive \cite{momeni_synthetic_2021}.
On the other hand, the radiologist has ability to look at the potential CMB from different perspectives and consult it with others, whereas the system does not. Therefore, providing additional information about gender, age, injury, angiography scans, etc. might also turn out beneficial \cite{kuijf_semi-automated_2013}.

When designing such a system from the clinical application point of view, certain practical aspects must also be taken into account.
It is important to remember about the end-user's perspective.
In this context, the form of results presentation should be designed considering user experience.

Obviously, the indication by a bounding-box or circle should be provided, but other useful information such as the confidence score of the prediction could also be included.
This value is rather provided by the machine-learning system, but it gives the information to the radiologist about certainty, which can accelerate the rating process.

Other idea might be the presentation of results based on the existing rating scales such as MARS \cite{gregoire_microbleed_2009} or BOMBS \cite{Cordonnier2009} (Section \ref{sec:scales}).

Moreover, there should be the ability of result acceptance or rejection.
As the system to be designed is the \textit{computer aided} system, the user should have the possibility to agree with the proposed result or not, as his decision is final.
This decision, however, has to be strongly distinguished from involving a human in the loop.
The raters have knowledge essential for CMBs rating, therefore they may be used during system design, for instance to validate preliminary results or to label extracted candidates as CMB and non-CMB, similarly as in \cite{chen_toward_2019}, but they should not be used as the last stage of the process to increase the system performance.
The reported 100\% precision or specificity of a semi-automated system in which a human is part of the FP reduction process is simply misleading \cite{barnes_semiautomated_2011, kuijf_semi-automated_2013, morrison_user-guided_2018}.
Even if it significantly reduces the single scan rating time, this type of evaluation is confusing.

However, the feedback from the radiologist about the prediction may be used for continual learning \cite{pianykh_continuous_2020}. This kind of approach may cause improvement of the already working system.

This smoothly leads to the problem of system evaluation. Section \ref{sec:evaluation} presented different metrics and their correlations.
Depending on the selected task: classification, detection, or segmentation - different metrics are used.
However, it is crucial to present as many metrics as possible, as they focus on different aspects of the system.
The sensitivity of 99\% may seem an outstanding result, but when it goes with precision of 40\% it is not satisfying.
The researchers sometimes stress out the importance of sensitivity and diminish the number of potential false positives, but it can be harmful in terms of reliable system synthesis \cite{barnes_semiautomated_2011}.
All this leads to the conclusion that a properly designed system should be balanced and optimized as a whole.

System evaluation is also important for enabling comparison between different approaches.
It is clearly visible in Table \ref{approaches_comparison_table} that the researchers not always provide all necessary metrics, which significantly hinders identifying the state-of-the-art.

Despite the current levels of metrics, there is a general issue of system trustworthiness.
While the systems based on morphological operations and traditional image transformations are pretty easy to explain, the interpretability of black-box machine-learning systems is still a challenging task \cite{the_precise4q_consortium_explainability_2020, barredo_arrieta_explainable_2020, angelov_explainable_2021}.
This problem is crucial, especially in such a life-impacting domain as medicine.
The process of decision making should be clear to ensure that the conclusions are drawn based on the nature of the examined object and not on the bias. Therefore, there is an urgent need of bias reduction \cite{mikolajczyk_towards_2021}.
A list of guidelines regarding designing a responsible and trustworthy AI system is given in \cite{GoogleAI}.

To the best of our knowledge, the paper collates all available research reports regarding automatic cerebral microbleeds detection.
The challenges of this task and some flaws of existing proposals have been outlined.
We believe that this paper will serve as a mine of knowledge and ideas for further research within this domain.
We hope that it will stimulate better practices regarding exchanging knowledge between different research groups.

\end{document}